\documentclass[10pt,twocolumn,letterpaper]{article}

\usepackage{iccv}
\usepackage{times}
\usepackage{epsfig}
\usepackage{graphicx}
\usepackage{amsmath}
\usepackage{amssymb}

\usepackage{makecell}

\newcommand{\norm}[1]{\left\lVert#1\right\rVert}

\usepackage{multirow}
\usepackage{subcaption}

\usepackage[symbol]{footmisc}

\usepackage{mathtools}
\usepackage[ruled, linesnumbered]{algorithm2e}
\SetAlFnt{\small}
\SetAlCapFnt{\small}
\SetAlCapNameFnt{\small}

\usepackage[table,x11names, dvipsnames]{xcolor}

\usepackage{multicol}
\usepackage{lipsum}
\usepackage{mwe}

\definecolor{mygray}{gray}{0.9}

\usepackage{array}
\newcolumntype{L}[1]{>{\raggedright\let\newline\\\arraybackslash\hspace{0pt}}m{#1}}
\newcolumntype{C}[1]{>{\centering\let\newline\\\arraybackslash\hspace{0pt}}m{#1}}
\newcolumntype{R}[1]{>{\raggedleft\let\newline\\\arraybackslash\hspace{0pt}}m{#1}}
\newcommand*{\affmark}[1][*]{\textsuperscript{#1}}

\newlength\mylen

\usepackage[pagebackref=true,breaklinks=true,letterpaper=true,colorlinks,bookmarks=false]{hyperref}
\iccvfinalcopy % *** Uncomment this line for the final submission

\def\iccvPaperID{5417} % *** Enter the ICCV Paper ID here

\ificcvfinal\fi
\begin{document}

%%%%%%%%% TITLE
\title{Hyperpixel Flow: Semantic Correspondence with Multi-layer Neural Features}%:\\ Fast and Accurate Image Matching with Selected Hypercolumns}

% \author[1]{Juhong Min}
% \author[1]{Alice Smith}
% \author[2]{Bob Jones}
% \affil[1]{Department of Mathematics, University X}
% \affil[2]{Department of Biology, University Y}

% \author{Juhong Min\affmark[\dag]\hspace{0.8cm}Jongmin Lee\affmark[\dag]\hspace{0.8cm}Jean Ponce\affmark[\S]\affmark[*]\hspace{0.8cm}Minsu Cho\affmark[\dag]\vspace{1.5mm}\\
% \affmark[\dag]POSTECH \hspace{1.5cm} \affmark[\S]DI ENS\vspace{3.0mm} \hspace{1.5cm} \affmark[*]Inria \\
% % \bf{-- This is a draft for an internal review. Do not distribute. --}\\
% }

\author{Juhong Min\affmark[1,2]\hspace{0.8cm}Jongmin Lee\affmark[1,2]\hspace{0.8cm}Jean Ponce\affmark[3,4]\hspace{0.8cm}Minsu Cho\affmark[1,2]\vspace{1.5mm}\\
\affmark[1]POSTECH \hspace{1.5cm} 
\affmark[2]NPRC\footnotemark[1] \hspace{1.5cm}  
\affmark[3]Inria\vspace{3.0mm} \hspace{1.5cm} 
\affmark[4]DI ENS\footnotemark[2] \\
{\tt\small \url{http://cvlab.postech.ac.kr/research/HPF/}}
}

\maketitle

\newcommand{\mcho}[1]{\textcolor{magenta}{#1}}
\newcommand{\jmin}[1]{\textcolor{blue}{#1}}
\newcommand{\jmlee}[1]{\textcolor{cyan}{#1}}

% !TEX root = ../main.tex

%%%%%%%%% 0. ABSTRACT
\begin{abstract}
Establishing visual correspondences under large intra-class variations %, \eg, between different instances of the same object category. 
%What makes the task so challenging? 
%requires discriminating different parts of images by 
requires analyzing images at different levels, from features linked to semantics and context to local patterns, while being invariant to instance-specific details.
To tackle these challenges, we represent images by ``hyperpixels'' that leverage a small number of relevant features selected among early to late layers of a convolutional neural network. %The feature layers are selected  selection requires only a small validation set of supervised image pairs. 
Taking advantage of the condensed features of hyperpixels, we develop an effective real-time matching algorithm based on Hough geometric voting. 
The proposed method, hyperpixel flow, sets a new state of the art on three standard benchmarks as well as a new dataset, SPair-71k, which contains a significantly larger number of image pairs than existing datasets, with more accurate and richer annotations for in-depth analysis. %, proving the robustness of our approach.
\end{abstract}

\footnotetext[1]{The Neural Processing Research Center, Seoul, Korea}
\footnotetext[2]{D\'epartement d'informatique de l'ENS, ENS, CNRS, PSL University, Paris, France}

% !TEX root = ../main.tex

%%%%%%%%% 1. Introduction
\section{Introduction}
Establishing visual correspondences under large intra-class variations, \ie, matching scenes depicting different instances of the same object categories, remains a challenging problem in computer vision. %This is an essential problem with a wide range of applications such as co-localization~\cite{cho2015unsupervised}, object tracking, stereo-matching, image retrieval and object re-identification. Despite its importance and applicability, 
%It is a highly challenging problem due to large intra-class, view-point and scale variations and deformations. 
It requires analyzing scenes at different levels, from features linked to semantics and context to local image patterns, while being invariant to irrelevant instance-specific details.
Recent methods have addressed this problem using deep convolutional features. 
Many of them~\cite{choy2016universal,han2017scnet,kim2017fcss,Rocco18} formulate this task as local region matching and learn to assign a local region in an image to a correct match in another image. %because the deep features encode high-level semantics with their receptive fields covering the entire input images, the correlations between region pairs can hardly be characterized, causing local-ambiguity.
Others~\cite{NIPS2018_7851, Rocco17, Rocco18, paul2018attentive} cast it as image alignment and learn to regress the parameters of global geometric transformation, \eg, using an affine or thin plate spline model~\cite{donato2002approximate}. %, aligning an image with another. 
These methods, however, mainly perform the prediction based on the output of the last convolutional layer, and fail to fully exploit the different levels of semantic features available to resolve the severe ambiguities in matching linked with intra-class variations.  
% \begin{figure}[t]
% 	\begin{subfigure}{0.5\textwidth}
% 	    \centering
%         \includegraphics[width=0.9\linewidth]{figures/Figure1.pdf}%\vspace{-2.5mm}
% 		%\caption{Hyperpixel}
% 	\end{subfigure} 
% 	\begin{subfigure}{0.5\textwidth} 
% 	    \centering\vspace{1mm}
%         \includegraphics[width=0.9\linewidth]{figures/Figure2.pdf}%\vspace{-1.5mm}
% 		%\caption{Hyperpixel flow}
% 	\end{subfigure}\vspace{-2mm}
% 	\caption{Hyperpixel flow. {Top:} The {\em hyperpixel} is a multi-layer pixel representation created with selected levels of features optimized for semantic correspondence. It provides multi-scale features, resolving local ambiguities. {Bottom:} The proposed method, {\em hyperpixel flow}, establishes dense correspondences in real time using hyperpixels. %More confident matches are colored in more yellow.
% 	} \label{fig:teaser}\vspace{-8.0mm}
% \end{figure}

\begin{figure}[t]
	
    \centering
    \includegraphics[width=1.0\linewidth]{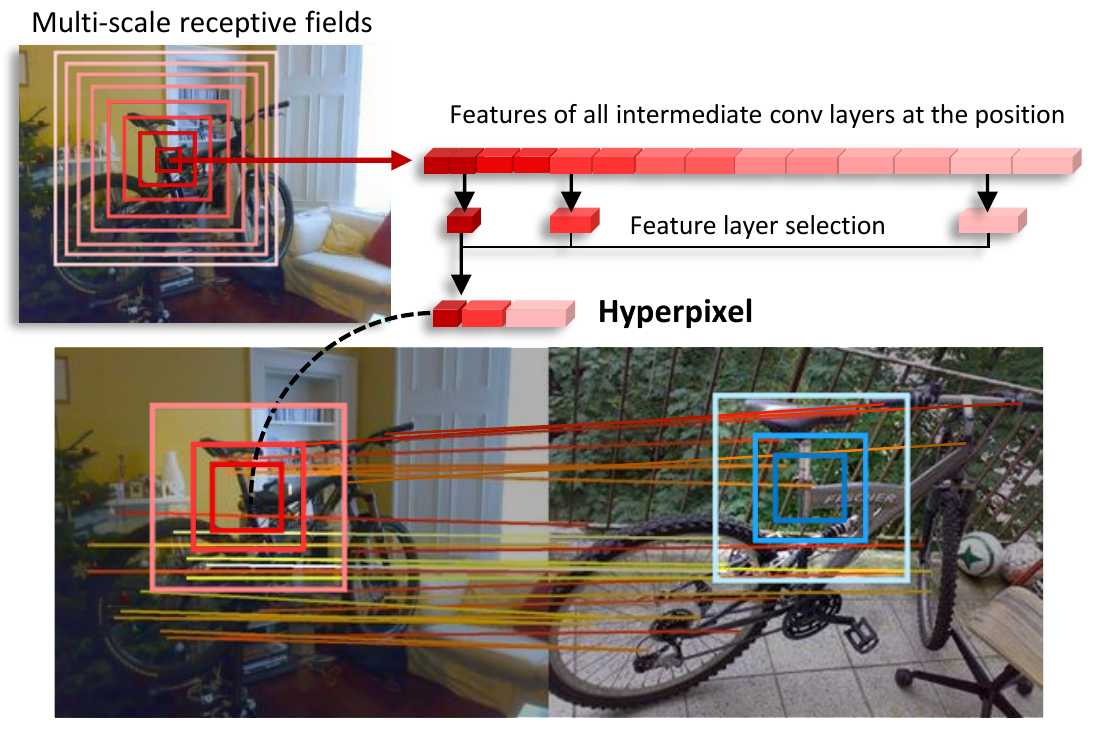}
	\vspace{-3.0mm}
		
	\caption{Hyperpixel flow. {Top:} The {\em hyperpixel} is a multi-layer pixel representation created with selected levels of features optimized for semantic correspondence. It provides multi-scale features, resolving local ambiguities. {Bottom:} The proposed method, {\em hyperpixel flow}, establishes dense correspondences in real time using hyperpixels. %More confident matches are colored in more yellow.
	} \label{fig:teaser}
	\vspace{-3.0mm}
\end{figure}

We propose a novel dense matching method, dubbed {\em hyperpixel flow} (Figure~\ref{fig:teaser}). Inspired by the hypercolumns~\cite{hariharan2015hypercolumns} used in object segmentation and detection, we represent images by ``hyperpixels'' that leverage different levels of features among early to late layers of a convolutional neural network and disambiguate parts of images in multiple visual aspects. The corresponding feature layers for hyperpixels are selected by a simple yet effective search process which requires only a small validation set of supervised image pairs.  
We show that the resultant hyperpixels provide both fine-grained and context-aware features suited for semantic correspondence and that only a few layers are sufficient and even better for the purpose, thus making hyperpixels an effective representation for light-weight computation. To obtain a geometrically consistent flow of hyperpixels, we present a real-time dense matching algorithm, regularized Hough matching (RHM), building on a recent region matching method using geometric voting~\cite{cho2015unsupervised}. 
Furthermore, we also introduce a new large-scale dataset, SPair-71k, with more accurate and richer annotations, which facilitates in-depth analysis for semantic correspondence.

%The proposed method, {\em hyperpixel flow}, sets new state-of-the-art results on three standard benchmarks as well as our new dataset, SPair-71k. 
%In addition, we introduce a new benchmark dataset that contains richer annotations and significantly larger number of image pairs than the existing datasets do. With our underlying philosophy that the simplest solution tends to be the best one, state-of-the-art results are demonstrated on standard benchmarks of PF-WILLOW, PF-PASCAL, Caltech-101 and our new benchmark dataset as well. 

Our paper makes four main contributions:\vspace{-5px}
\begin{itemize}
\setlength\itemsep{0em}
    \item We propose {\em hyperpixels} for establishing reliable dense correspondences between two images, which provide multi-layer features robust to local ambiguities.
    \item We present an efficient matching algorithm, regularized Hough matching (RHM), that achieves a speed of more than 50 fps on a GPU for $300\times200$ image pairs.
    \item We introduce a new dataset, {\em SPair-71k}, which contains a significantly larger number of image pairs with richer annotations than existing ones.
    \item The proposed method, {\em hyperpixel flow}, sets a new state of the art on standard benchmarks as well as SPair-71k.
\end{itemize}
% !TEX root = ../main.tex
%%%%%%%%% 2. Related Work

\begin{figure*}
    \begin{center}
        \includegraphics[width=1.0\linewidth]{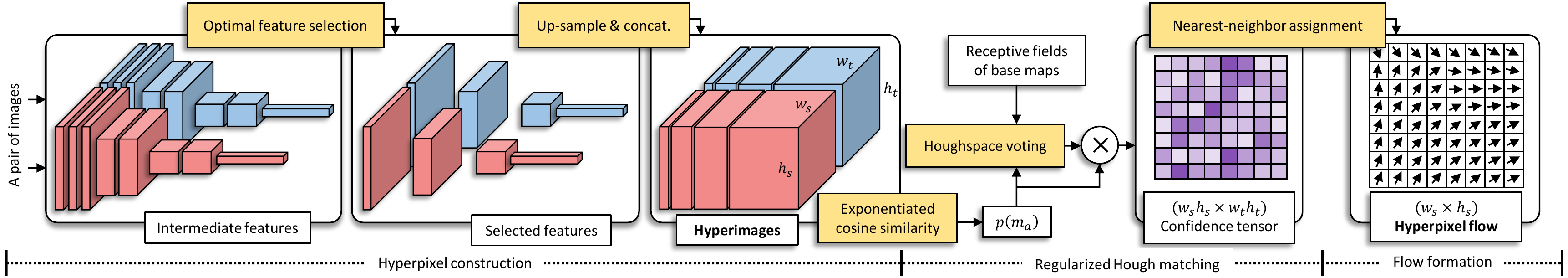}
    \end{center}
    \vspace{-2.0mm} 
       \caption{Overall architecture of the proposed method. Hyperpixel flow consists of three main steps: hyperpixel construction, regularized Hough matching, and flow formation. For details, see text.}
    \vspace{-3.0mm} 
\label{fig:architecture}
\end{figure*}

\section{Related Work}
\smallbreak
\noindent \textbf{Local region matching.}
Early methods commonly tackle semantic correspondence by matching two sets of local regions based on handcrafted features. %Lowe~\cite{lowe2004distinctive} proposes SIFT to solve correspondence problem. 
%Barnes \etal.~\cite{barnes2009patchmatch} propose the patch-match algorithm which find a patch correspondence to define nearest-neighbor field using random search. 
Liu~\etal~\cite{liu2016sift} and Kim~\etal~\cite{kim2013deformable} use dense SIFT descriptors to establish a flow of local regions across similar but different scenes by leveraging a hierarchical optimization technique in a coarse-to-fine manner. %Kim~\etal~\cite{kim2013deformable} propose a deformable spatial pyramid model (DSP) to find dense correspondences more efficiently. 
Bristow~\etal~\cite{bristow2015dense} use LDA-whitened SIFT descriptors, making correspondence more robust to background clutter.
Cho~\etal~\cite{cho2015unsupervised} introduce an effective voting-based algorithm based on region proposals and HOG features~\cite{dalal2005histograms} for semantic matching and object discovery. Ham~\etal~\cite{ham2016proposal} further extend the work with a local-offset matching algorithm, and introduce a benchmark dataset with keypoint-level annotations. %Their experiment shows that the HOG descriptor gives better matching performance than deep convolutional features. 
Taniai~\etal~\cite{taniai2016joint} tackle semantic correspondence jointly with cosegmentation, introducing a benchmark dataset annotated with dense flows and segmentation masks. 
All these hand-crafted representation fails to capture high-level semantics enough to discriminate complex patterns with large intra-class deformations. 

In this context, CNN features have emerged as good alternatives for semantic matching.
Long~\etal~\cite{long2014convnets} show that convolutional features from a CNN pretrained on classification are transferable to correspondence problems. 
Choy \etal~\cite{choy2016universal} attempt to learn a similarity metric based on a CNN using a contrastive loss with hard negative mining. %They compute patch similarity to train deep neural networks. %\jmlee{but it does not have good performance on recent benchmark.}
Han~\etal~\cite{han2017scnet} propose to learn a CNN end-to-end with geometric matching, which uses region proposals as matching primitives. Kim~\etal~\cite{kim2017fcss} introduce a CNN-based self-similarity feature for semantic correspondence, and also use it to estimate dense affine-transformation fields by an iterative discrete-continuous optimization~\cite{kim2017dctm}.
Novotny~\etal~\cite{novotny2018self} train a geometry-aware feature in an unsupervised regime and use it for part matching and discovery by measuring confidence scores.
Rocco~\etal~\cite{rocco2018neighbourhood} propose a neighbourhood consensus network that computes robust matching similarity using 4D convolution filters.
%However, We did not simply apply the CNN feature in local region matching, but we increased the performance by fully utilizing the hierarchical features that the CNN model can use.
% Although those methods may have succeeded in using the CNN feature, there was a drawback that the inference time was too long.
% Therefore, We increased the representation power by fully utilizing the hierarchical features available to the CNN model.

\smallbreak
\noindent \textbf{Global image alignment.}
%Jaderberg \etal~\cite{jaderberg2015spatial} proposed spatial transformer networks (STNs) using a parameterized sampling grid to apply geometric transformations. This can be used not only in a feed-forward network, but also a loss function to find dense correspondences. 
Some methods have cast semantic correspondence as global alignment.  
%Several methods~\cite{detone2016deep, kanazawa2016warpnet, thewlis2017unsupervised} propose to predict geometric transformation parameters using a CNN.
% directly using a CNN-based regressor.  
%This method allows the feature representation to encode local transformation information when global transformation parameter estimates.
% The supervision of global transformation parameters is given by synthetic pair generation~\cite{, Rocco17} and Rocco~\etal~\cite{Rocco17} proposed the full architecture of semantic alignment.
Rocco \etal~\cite{Rocco17} propose a CNN architecture which takes a correlation tensor and directly predicts global transformation parameters for geometric matching.
Seo \etal~\cite{paul2018attentive} improve it using offset-aware correlation kernels with attention.
Rocco~\etal~\cite{Rocco18} develop a weakly-supervised learning framework using differentiable soft-inlier count loss function.
Jeon \etal~\cite{jeon2018parn} propose a pyramidal affine transformation regression network to compute the correspondence hierarchically from high-level semantics to pixel-level points.
Kim \etal~\cite{NIPS2018_7851} introduce a recurrent alignment network that performs iterative local transformations with a global constraint. %These image alignment methods using global geometric transformation parameter are not precise when looking for pixel-level correspondence. Furthermore, they are sensitive to view-point variations or deformations.

\smallbreak
\noindent \textbf{Multi-layer neural features.} 
% Finding good local descriptors and choosing appropriate representation affects the performance of matching.
%Standard neural networks produce a final prediction relying on features  
%Recently, several methods research 
%In the tasks that consider local parts, top layer of CNN is not the best choice of representation.
Hariharan~\etal~\cite{hariharan2015hypercolumns} have shown that {\em hypercolumns} that combine features from multiple layers of CNN, improve object detection, segmentation, and part labeling. Following this work, several methods~\cite{kong2016hypernet, lin2017feature} have used multi-layer neural features with additional modules on object detection task.
%The hypercolumn feature concatenates low-level local features with upsampled high-level semantic features.  
% Kong~\etal~\cite{kong2016hypernet} use the hypercolumn features for region proposals by aggregating hierarchical feature maps and compressing them into a uniform space.
% Lin~\etal~\cite{lin2017feature} design a feature pyramid network which combines bottom-up feature extraction with top-down lateral connection for robustness to scale-variation.
% Liu~\etal~\cite{liu2018receptive} suggest a receptive field block which mixes an inception block and atrous spatial pyramid pooling on top of single-shot multi-box detector (SSD).
% Liu~\etal~\cite{liu2018receptive} suggest a receptive field block which mixes an inception block and atrous spatial pyramid pooling to capture multi-scale information.
% Liu~\etal~\cite{liu2018receptive} suggest a receptive field block to capture multi-scale information.
Fathy~\etal~\cite{fathy2018hierarchical} propose coarse-to-fine stereo matching method that uses  multi-layer features in sequence. 
%Savinov~\etal~\cite{savinov2017matching} suggest to find matches by comparing the neural activation paths on a CNN by assuming that similar parts have a similar activation pattern through the network.
%In the object detection field, the hierarchical features were used in various ways, but not in the semantic correspondence.
In semantic correspondence, multi-layer neural features have rarely been explored despite its relevance. Novotny~\etal~\cite{novotny2017anchornet} use residual hypercolumn features to learn a set of diverse filters for object parts.
Ufer and Ommer~\cite{ufer2017deep} employ pyramids of pre-trained CNN features to localize salient feature points guided by object proposals, and match them across images using sparse graph matching. In these methods, multi-layer features are mainly used to localize salient parts and the feature layers are manually selected following previous methods~\cite{girshick2015deformable, he2016deep}. 
Unlike these approaches and the hypercolumn~\cite{hariharan2015hypercolumns}, we use a multi-layer neural feature as a pixel representation for dense matching and optimize feature layers via layer search for the purpose.
We show that specific combinations of layers significantly affect matching performance and using only a small number of layers can achieve a remarkable performance. %Unlike existing hypercolumn, Hyperpixel representation has each spatial index with its value for the feature and its receptive field information. 

\smallbreak
\noindent \textbf{Neural architecture search (NAS).} 
The layer search for hyperpixels can be viewed as an instance of NAS~\cite{liu2018darts, randomwire, zoph2016neural, Zoph_2018_CVPR}. %in the context of searching for an optimal layer combination. 
Unlike a general search space of network configurations in NAS, however, the search space in our work is limited to combinations of feature layers for visual correspondence.
\vspace{-2.0mm}
% \mcho{These kinds of method has been successful to use deep neural features for semantic correspondence, but CNN features can be used to discriminate semantically similar part of each object more robust by simple way. (revise and clarify.)}

% !TEX root = ../main.tex

%%%%%%%%% 3. Hyperpixel Flow
\section{Hyperpixel Flow}

Our method presented below, dubbed {\em hyperpixel flow}, can be divided into three steps: (1) hyperpixel construction, (2) regularized Hough matching, and (3) flow formation. Figure~\ref{fig:architecture} illustrates the overall architecture of our model aligned with the three steps. Each input image is fed into a convolutional neural network to create a set of hyperpixels. The hyperpixels are then used as primitives for the regularized Hough matching algorithm to build a tensor of matching confidences for all candidate correspondences. 
The confidence tensor is transformed into a hyperpixel flow in a post-processing step assigning a match to each hyperpixel.  
Three steps are detailed in this section. 

\subsection{Hyperpixel construction}
Given an image, a convolutional neural network produces a sequence of $L$ feature maps $(\mathbf{f}^0, \mathbf{f}^1, ..., \mathbf{f}^{L-1})$ as intermediate outputs. We represent the original image by a {\em hyperimage} by pooling {\em a small} subset of $K$ feature maps, optimized for semantic correspondence, and concatenating them along channels with upsampling: 
\begin{align}
    \mathbf{F} = \big[ \mathbf{f}^{l_0}, \zeta(\mathbf{f}^{l_1}), \zeta(\mathbf{f}^{l_2}), ..., \zeta(\mathbf{f}^{l_{K-1}}) \big],
\end{align}
where $\zeta$ denotes a function that upsamples the input feature map to the size of $\mathbf{f}^{l_0}$, the {\em base} map. 
We can associate with each spatial position $p$ of the hyperimage the corresponding image coordinates, a hyperpixel feature, and its multi-scale receptive fields.
%\begin{align}
%    \mathcal{H} = (\mathbf{h}_1, \mathbf{h}_2, ..., \mathbf{h}_N),
%\end{align}
%where $p$ denotes the spatial position of a particular geometry, $1 \leq p \leq N$. Each vector in hyperfeature, $h_k$, has a bijective relationship with a particular receptive field, $r_k$. 
Let us denote by $\mathbf{x}_p$ the image coordinate of position $p$, and by  $\mathbf{f}_p$ the corresponding hyperfeature, \ie, $\mathbf{f}_p = \mathbf{F}({\mathbf{x}_p})$.
The hyperpixel at position $p$ on the hyperimage is then defined as   
\begin{align}
   \mathbf{h}_p = (\mathbf{x}_p, \mathbf{f}_p).
\end{align}
As will be seen in the next subsection, the hyperpixels are used as primitives for the subsequent matching process. 
%\mcho{Add illustration pic for hyperpixels?}

\begin{algorithm}[t]
    %\DontPrintSemicolon
    \SetKwInOut{Input}{Input}  
    %\KwIn{$\mathcal{L}_{\textrm{cand}}$: candidate layers of a CNN}
    \Input{$\mathcal{L}_{\textrm{cand}} = \{0, ..., L-1\}$: all candidate layers\\
           $\mathcal{L}_{\textrm{base}}$:  candidate layers for the base ($\subset\mathcal{L}_{\textrm{cand}}$)\\%, 
           $N_{\textrm{beam}}$: the beam size\\
           $K_{\textrm{max}}$: the maximum number of layers allowed
           }
    \KwOut{$\mathcal{L}_\textrm{sel}$: the set of selected layers}
    \SetKwBlock{Begin}{function}{end function}
    \Begin($\text{SearchLayers}$)
    {
    		%\text{Initialize memory buffers $\mathcal{M}$ and $\mathcal{M}'$}\;
		\tcp{initialize memory buffers}
		$\mathcal{M}.\mathrm{init}();$ \ \ \ $\mathcal{M}'.\mathrm{init}()$\; 
		\tcp{base layer search}
        \ForAll{$l \in \mathcal{L}_{\mathrm{base}}$}
        {
            $v \gets$ evaluateLayerSet$(\{l\})$\;
            $\mathcal{M}$.insert($(\{l\},v)$)\;
        }
        
        $\mathcal{M}' \gets \mathcal{M}$.findBestN($N_{\mathrm{beam}}$)\; 
        $({\mathcal{L}_\mathrm{sel}}, v_\mathrm{sel}) \gets \mathcal{M}'$.findBest()\; 
        
        \tcp{layer search iterations}    
        \For{$k \gets 1$ to $K_\mathrm{max}-1$}
        {
            $\mathcal{M}.\mathrm{init}()$\;  
            \ForAll{$({\mathcal{L}'},v') \in \mathcal{M}'$}
            {
                \ForAll{$l \in \mathcal{L_{\mathrm{cand}}}$}
                {
                    \If{$l \notin \mathcal{L}' \land l > \mathrm{min}(\mathcal{L}')$}
                    {
                        $v'' \gets$ evaluateLayerSet$(\mathcal{L}' \cup \{l\})$\;
                        $\mathcal{M}$.insert($\mathcal{L}' \cup \{l\}, v'')$)\;
                    }
                }
            }
            $\mathcal{M}' \gets \mathcal{M}$.findBestN($N_{\mathrm{beam}}$)\; 
            $({\mathcal{L}^*}, v^*) \gets \mathcal{M}'$.findBest()\;
             \If{$v^* > v_\mathrm{sel}$}
            {
                $({\mathcal{L}_\mathrm{sel}}, v_\mathrm{sel}) \gets ({\mathcal{L}^*}, v^*)$\;
            }
        }
        
        \Return{$\mathcal{L}_\mathrm{sel}$}
    }\vspace{-1.0mm}
\caption{Beam search for hyperpixel layers.}
\label{alg:layersearch}
\end{algorithm} 

%\smallbreak
%\noindent \textbf{Hyperpixel layer search.} 
To select the optimal set of feature maps for hyperpixels, %While the work of \cite{long2014convnets} shows convolutional features are good at localizing at a much finer scale than their receptive field sizes, finding layers suited for correspondence still remains a challenging issue due to uninterpretable nature of deep features. 
%We take a convolutional network pretrained on classification, put all convolutional layers into a set of candidate layers, and 
%select a subset from it them using a validation dataset with ground-truth annotation. 
we perform a search over all convolutional layers of a given CNN so that a subsequent matching algorithm achieves the best validation performance. In our case, we use regularized Hough matching (Sec.~\ref{sec:RHM}) for the matching algorithm and the probability of correct keypoints (PCK) (Sec.~\ref{sec:PCK}) for the performance metric. 
For the search algorithm, we use a variant of beam search~\cite{beamsearch}, which is a breadth-first search algorithm with a limited memory. 
Basically, at each iteration, it evaluates the effect of each candidate layer by adding it to current combinations of layers in the memory and then replaces them with a fixed number of top performing combinations.  
%$N_{\mathrm{beam}}$ best performing base layers are collected based on their performance on a validation set and stored in a memory buffer $\mathcal{M}'$ for the subsequent layer search process; similarly, among all possible layer combinations obtained by appending a candidate layer $l \in \mathcal{L}_{\mathrm{cand}}$ to each combination in $\mathcal{M}'$, $N_{\mathrm{beam}}$ best performing layer combinations are collected and stored in $\mathcal{M}'$.
The search process is repeated until the number of selected layers reaches the maximum number of layers allowed. Finally, we use the best combination found along the search. 
The detailed procedure is summarized in Algorithm~\ref{alg:layersearch}, where we restrict base layer candidates, $\mathcal{L}_{\mathrm{base}}$ only to layers with a sufficient spatial resolution. 
%By taking the advantage of fast inference time of our model, we adopted greedy search over the intermediate layer space.
%each candidate is drawn to get evaluated on validation set and is appended to a list of approved layers if and only if the evaluation result on the approved layers with drawn candidate layer is better than the result on previously approved layers. This greedy procedure is done in recursive manner, adding one candidate after another to the approved layer. The detailed procedure is described in Algorithm~\ref{alg:algorithm}. 
%\mcho{:the algorithm part needs to be revised.}

\subsection{Regularized Hough matching}\label{sec:RHM}
In order to establish visual correspondences, we adapt the probabilistic Hough matching (PHM), algorithm of Cho \etal ~\cite{cho2015unsupervised}, to hyperpixels. The key idea of PHM is to re-weight appearance similarity by Hough space voting to enforce geometric consistency. In our context, let $\mathcal{D}=(\mathcal{H}, \mathcal{H}')$ be two sets of hyperpixels, and $m=(\mathbf{h},\mathbf{h}')$ be a hyperpixel match where $\mathbf{h}$ and $\mathbf{h}'$ are respectively elements of $\mathcal{H}$ and $\mathcal{H}'$. Given a Hough space $\mathcal{X}$ of possible offsets (image transformation) between the two hyperpixels,  
the confidence for match $m$, $p(m|\mathcal{D})$, is computed as  
\begin{align}
    \label{eqn:houghmatching} 
    p(m|\mathcal{D}) &\propto  p(m_\mathrm{a})\sum_{\mathbf{x}\in \mathcal{X}}p(m_\mathrm{g}|\mathbf{x})\sum_{m \in \mathcal{H} \times \mathcal{H}'}p(m_\mathrm{a})p(m_\mathrm{g}|\mathbf{x}),  %\\
   % &= p(m_{a})p(m_{g}),
\end{align}
where $p(m_\mathrm{a})$ represents the confidence for appearance matching and $p(m_\mathrm{g}|\mathbf{x})$ is the confidence for geometric matching with an offset $\mathbf{x}$, measuring how close the offset induced by $m$ is to $\mathbf{x}$. % implemented by a discretized Gaussian kernel centered on $\mathbf{x}$. 
By sharing the Hough space $\mathcal{X}$ for all matches, PHM efficiently computes the match confidence with good empirical performance~\cite{cho2015unsupervised,ham2016proposal,han2017scnet}.

%\smallbreak
%\noindent\textbf{Semantic appearance similarity.} 
In this work we compute appearance matching confidence using hyperpixel features:    
%Due to the nature of CNNs, lower layer features have relatively fewer channels than deeper layer features do, meaning a large part of the hyperpixels hugely depends upon the general semantics of an object, rather than its parts, causing ambiguation between hyperpixel semantics. As we  need a way to discriminate one hyperpixel semantics to the other, we modify the appearance matching term in~\ref{eqn:houghmatching} as below.
 \begin{align}
    p(m_\mathrm{a})
    = \mathrm{ReLU}\Big( \frac{\mathbf{f} \cdot \mathbf{f}'}{\norm{\mathbf{f}} \norm{\mathbf{f}'}} \Big)^d,
\end{align}
where 
%; the pair already far away from each other will get much further away than the pair relatively closer than the former, discriminating semantics. 
the ReLU function clamps negative values to zero and the exponent $d$ is used to emphasize the difference between the hyperpixel features.
When combined with Hough voting, this similarity function with $d \geq 2$ improves matching performance by suppressing noisy activations. We set $d=3$ in our experiments.

%\smallbreak
%\noindent\textbf{Hough space geometric voting.}
%For the sake of brevity, we use short notations $\mathcal{T}_{s}$ and $\mathcal{T}_{t}$ for affine transformation matrices and $v_{s}$ and $v_{t}$ for the reference vectors. Applying $\mathcal{T}_{s}^{-1}$ and $\mathcal{T}_{t}$ consecutively to the source reference vector $v_{s}$ yields a vector that approximates Hough space bin index for $m$, denoted as $\psi(m)$.
%\begin{align}
%    \psi(m) = \mathcal{T}_{t}\mathcal{T}_{s}^{-1}v_{s}.
%\end{align}
To compute $p(m_\mathrm{g}|\mathbf{x})$, we construct a two-dimensional offset space, quantize it into a grid of bins, and use a set of center points of the bins for $\mathcal{X}$. For Hough voting, each match $m$ is assigned to the corresponding offset bin to increment the score of the bin by the appearance similarity score, $p(m_{\mathrm{a}})$. Despite their (serial) complexity of $O(|\mathcal{H}| \times |\mathcal{H}'|)$, the operations are mutually independent, and can thus easily be parallelized on a GPU. %For additional details, please refer to our project website. % the supplementary material. 
%\mcho{Describe how this computaion is parallelized for GPU in our implementation.}

Previous versions of PHM all use multi-scale region proposals~\cite{manen2013prime, pont2017multiscale, uijlings2013selective} as matching primitives described with HOG~\cite{cho2015unsupervised,ham2016proposal} or a single feature map from a CNN~\cite{ham2016proposal,han2017scnet}. While using irregular and multi-scale region proposals focuses attention on object-like regions, it requires creating a three-dimensional offset space for translation and scale changes with higher memory and computation. 
%To find corresponding offset bins for Hough voting, the height and width of the boxes are treated as scaling factors while the center of the boxes are used as translation factors of affine transformations. The width and height of the input images play a role of reference vector of the transformations as mere transformation parameters represent nothing but only a transformation. 
%Since the original work~\cite{cho2015unsupervised} uses bottom-up region proposal method which produces boxes varying in sizes and aspect ratios, the dimension for change in aspect ratios in Hough space is necessary. 
In contrast, the use of hyperpixels reduces the Hough space down to two dimensions and makes the voting procedure faster and simpler since all hyperpixels are homogeneous on a predefined regular grid.
In addition, unlike region proposals, hyperpixels provide (quasi-)dense image features and their multi-layer features improve performance in practice. 
%Furthermore, the dense hyperpixels mitigate the danger of missing parts in sparse hard attention of proposals and improve the matching performance with their rich multi-layer features. 
% changes in aspect ratio always equal to one for hyperpixels and their receptive fields. 
%Also, with prior knowledge that the width and height of the hyperpixels are all the same for both $\mathcal{I}_s$ and $\mathcal{I}_t$, we further optimize this voting scheme.
%If bottom-up region proposal methods are used, acquiring boxes with extreme scales and aspect ratios is inevitable. Thus, process of pruning boxes that are unnatural for matching is necessary. In case of receptive fields, however, the model knows about the lower and upper bound of $\psi(m)$ in prior because the centers of the hyperpixels are bounded between input image sides: $0 \leq t_{x,s} \leq w_{s}$, $0 \leq t_{x,t} \leq w_{t}$, $0 \leq t_{y,s} \leq h_{s}$, and $0 \leq t_{y,t} \leq h_{t}$. The lower and upper bound for $\alpha$ and $\beta$ is given by
%\begin{align}
%    0 \leq \alpha \leq w_{s} + w_{t}, \quad
%    0 \leq \beta \leq h_{s} + h_{t},
%\end{align}
%which allows the model to remove the existing process of pruning boxes with extreme sizes and aspect ratios.
%Using hyperpixels reduces the houghspace dimension and exposes a large number of dispensable computations. 
In our GPU implementation, our algorithm, {\em regularized Hough matching} (RHM), runs 100 to 500 times faster than PHM (2$\sim$20 msecs vs. 1$\sim$2 secs), enabling real-time matching.
\iffalse
\begin{align}
    \psi(m) &= 
    \begin{bmatrix}
        s & 0    & t_{x,s}\\
        0    & s & t_{y,s} \\
        0    & 0       & 1
    \end{bmatrix}
    \begin{bmatrix}
        s & 0    & t_{x,t}\\
        0    & s & t_{y,t} \\
        0    & 0       & 1
    \end{bmatrix}^{-1}
    \begin{bmatrix}
        w_{s} \\
        h_{s} \\
        1
    \end{bmatrix} \\
    &= 
    \begin{bmatrix}
        w_{s} - t_{x,s} + t_{x,t}\\
        h_{s} - t_{y,s} + t_{y,t}\\
        1
    \end{bmatrix} = 
    \begin{bmatrix}
        \alpha\\
        \beta\\
        1
    \end{bmatrix}.
\end{align}
\fi

\begin{center}
\begin{table*}
    \centering
    \scalebox{0.72}{
    \begin{tabular}{C{3.15cm}|C{1.65cm}|c|C{3.5cm}|C{4.2cm}|C{4cm}|C{3cm}}%{p{2.75cm}|| p{0.75cm}| p{0.75cm}| p{0.75cm}| p{3cm}| p{3.25cm} | p{3cm} }
        \hline
        Dataset name & Size (pairs)  & Class &  Source datasets & Annotations & Characteristics & Users of the dataset\\
        \hline\hline
         Caltech-101~\cite{kim2013deformable} & 1,515  &  101  &  Caltech-101~\cite{fei2004learning,li2006one}  & object segmentation &  tightly cropped images of objects, little background & ~\cite{ham2016proposal, han2017scnet, jeon2018parn, kim2017fcss, laskar2019semantic, Rocco17, Rocco18, paul2018attentive}\\
         \hline
         PASCAL-PARTS~\cite{zhou2015flowweb} & 3,884  &  20  &      PASCAL-PARTS~\cite{chen_cvpr14}, PASCAL3D+~\cite{Xiang2014BeyondPA} & keypoints (0$\sim$12), azimuth, elevation, cyclo-rotation, body part segmentation & tightly cropped images of objects, little background, part and 3D infomation & ~\cite{choy2016universal,han2017scnet,kim2017fcss,kim2017dctm,novotny2017anchornet,ufer2017deep} \\
         \hline
         Animal-parts~\cite{novotny16i-have} &  $\approx$7,000 & 100 &   ILSVRC 2012 ~\cite{krizhevsky2012imagenet} &  keypoints (1$\sim$6)  & keypoints limited to eyes and feet of animals & ~\cite{novotny2017anchornet} \\
         \hline
         CUB-200-2011~\cite{WahCUB_200_2011} & 120k & 200 & CUB-200-2011~\cite{WahCUB_200_2011}  & 15 part locations, 312 binary attributes, bbox & tightly cropped images of object, only bird images & ~\cite{choy2016universal, kanazawa2016warpnet}  \\

         \hline
         TSS~\cite{taniai2016joint} &  400 &   9  & FG3DCar~\cite{lin2014jointly}, JODS \cite{rubinstein2013unsupervised}, PASCAL~\cite{hariharan2011semantic} &  object segmentation, flow vectors &  cropped images of objects, moderate background & ~\cite{cho2015unsupervised,ham2016proposal,jeon2018parn,NIPS2018_7851,kim2017fcss,kim2017dctm,laskar2019semantic,Rocco17,Rocco18,paul2018attentive}  \\
         \hline
         PF-WILLOW~\cite{ham2016proposal} &   900 & 5 &  PASCAL VOC 2007~\cite{everingham2015pascal}, Caltech-256~\cite{cho2013learning, griffin2007caltech} &  keypoints (10) & center-aligned images, pairs with the same viewpoint & ~\cite{ham2016proposal,han2017scnet,NIPS2018_7851,kim2017fcss,kim2017dctm,novotny2017anchornet,Rocco17,paul2018attentive,ufer2017deep} \\
         \hline
         PF-PASCAL~\cite{ham2018proposal}  &  1,300  & 20 & PASCAL VOC 2007~\cite{everingham2015pascal} &  keypoints (4$\sim$17), bbox. & pairs with the same viewpoint & ~\cite{ham2016proposal,han2017scnet,jeon2018parn,NIPS2018_7851,laskar2019semantic,novotny2018self,Rocco17,Rocco18,rocco2018neighbourhood,paul2018attentive} \\
         \hline\hline
         SPair-71k (ours)  &   70,958 & 18 & PASCAL3D+~\cite{Xiang2014BeyondPA}, PASCAL VOC 2012~\cite{everingham2015pascal} & keypoints (3$\sim$30), azimuth, view-point diff., scale diff., trunc. diff., occl. diff., object seg., bbox. & large-scale data with diverse variations, rich annotations, clear dataset splits &  this work\\
         \hline
    \end{tabular} 
    }
    \vspace{-3.0mm} 
    \caption{Public benchmark datasets for semantic correspondence.  The datasets are listed in chronological order. Research papers using the datasets for evaluation are listed in the last column. See text for details.}
    \vspace{-5.0mm} 
    \label{tab:benchmark_survey}
\end{table*}
\end{center}

\vspace{-11.0mm}
\subsection{Flow formation and keypoint transfer}
The raw output of RHM is a tensor of confidences for all candidate matches. It can easily be transformed into a hyperpixel flow in a post-processing step of assigning a match to each hyperpixel, \eg, by nearest-neighbor assignment.
Since the base map of the hyperimage is selected among early layers, the flow is dense enough for many applications. 

Transferring keypoints from an image to the corresponding points in another image is commonly used for evaluating semantic correspondences.  
%To achieve this, building a pixel-level flow field is a required procedure for previous part-based approaches~\cite{ham2016proposal, han2017scnet}. Because they use object proposals as matching primitives which are not dense enough, comparing matching scores and applying a joint image filtering are necessary steps to build a dense flow field.
%Another big advantage of Hyperpixel Flow approach is the fact that the output flow is already dense enough. 
We use a simple method for keypoint transfer using hyperpixel flow; 
given a keypoint $\mathbf{x}_p$ in a source image, its neighbor hyperpixels $ \mathcal{N}(\mathbf{x}_p)$ are collected whose base map receptive fields cover the keypoint, and the displacement vectors from the centers of the base map receptive fields to the keypoint, denoted by $\{ \mathbf{d}(\mathbf{x}_q) \}_{\mathbf{x}_q\in\mathcal{N}(\mathbf{x}_p)}$, are computed. Given the hyperpixel flow $T$ of $\mathcal{N}(\mathbf{x}_p)$ predicted by our method, we apply the average of the displacements $\{ T(\mathbf{x}_q) + \mathbf{d}(\mathbf{x}_q) \}_{\mathbf{x}_q\in\mathcal{N}(\mathbf{x}_p)}$ to localize a corresponding keypoint in the target image.
\section{SPair-71k dataset}
With growing interest in semantic correspondence, several annotated benchmarks are now available. 
Some popular ones are summarized in Table~\ref{tab:benchmark_survey}.  
% \smallbreak
% \noindent \textbf{Proposal Flow (PF).} There are two main benchmark datasets proposed by PF~\cite{ham2016proposal}, PF-WILLOW and PF-PASCAL. PF-WILLOW and PF-PASCAL contain 900 and 1300 semantically related image pairs respectively. For comparable evaluation with other models~\cite{Rocco17, Rocco18, han2017scnet} where they split the dataset 700, 300 and 300 pairs for training, validation and test set respectively, we used the same split. The details of these datasets will be discussed in supplementary material.
% \smallbreak
% \noindent \textbf{Caltech-101.} With 101 object categories, the dataset consists of 1515 semantically related image pairs, 15 pairs for each of the 101 object categories of the dataset. For a pair of input images, ground-truth contour keypoints wrapping objects of interest are given as annotations. The contour keypoints are used to measure three main evaluation metrics. First, the label transfer accuracy (LT-ACC) measures the amount of overlapping area with emphasis on background class. Second, the intersection-over-union (IoU) also measures the overlap but with stronger emphasis on foreground class. Lastly, the localization error (LOC-ERR) measures the difference between ground-truth and predicted dense flow. As mentioned in~\cite{Rocco18}, LOC-ERR is not evaluated since the annotations for LOC-ERR in Caltech-101 is so unrealistic that measuring this evaluation metric has no meaning on this benchmark.
Due to the high expense of ground-truth annotations for semantic correspondence, early benchmarks~\cite{chen_cvpr14, kim2013deformable} only support indirect evaluation using a surrogate evaluation metric rather than direct matching accuracy.    
For example, the Caltech-101 dataset in~\cite{kim2013deformable} provides binary mask annotations of objects of interest for 1,515 pairs of images and the accuracy of mask transfer is evaluated as a rough approximation to that of matching. 
% The contour keypoints are used to measure three main evaluation metrics. First, the label transfer accuracy (LT-ACC) measures the amount of overlapping area with emphasis on background class. Second, the intersection-over-union (IoU) also measures the overlap but with stronger emphasis on foreground class. Lastly, the localization error (LOC-ERR), measures the difference between ground-truth and predicted dense flow, is not evaluated to this task following to ~\cite{Rocco18} because of unrealistic measure on this task.
% As mentioned in~\cite{Rocco18}, LOC-ERR is not evaluated since the annotations for LOC-ERR in Caltech-101 is so unrealistic that measuring this evaluation metric has no meaning on this benchmark.
Recently, Ham~\etal~\cite{ham2016proposal,ham2018proposal} and Taniai~\etal~\cite{taniai2016joint} have introduced datasets with ground-truth correspondences. Since then, PF-WILLOW~\cite{ham2016proposal} and PF-PASCAL~\cite{ham2018proposal} have been used for evaluation in many papers. They contain 900 and 1,300 image pairs, respectively, with keypoint annotations for semantic parts. 

All previous datasets, however, have several drawbacks: 
%\begin{itemize}
%\setlength\itemsep{0em}
%    \item The dataset split was not consistent.
%    \item The amount of training data was insufficient.
%    \item There were not enough annotation types.
%\end{itemize}
First, the amount of data is not sufficient to train and test a large model. %Due to the reason, recent methods have used several techniques on top of the datasets such as  self-supervised learning with synthetic pair~\cite{Rocco17}, data augmentation~\cite{Rocco18}, and semi-supervised learning with unlabeled pairs~\cite{laskar2019semantic}.  
Second, image pairs do not display much variability in viewpoint, scale, occlusion, and truncation. 
%This makes the datasets relatively easy and inadequate to simulate realistic challenges.  
Third, the annotations are often limited to either keypoints or object segmentation masks, which hinders in-depth analysis.
%In case of PF-WILLOW, due to the lack of annotation type, PCK thresholding was performed by calculating the size of the object using the maximum distance of the keypoints. This is an inaccurate measurement.
Fourth, the datasets have no clear splits for training, validation, and testing. Due to this, recent evaluations in~\cite{han2017scnet, Rocco18, rocco2018neighbourhood} have been done with different dataset splits of PF-PASCAL. Furthermore, the splits are disjoint in terms of image pairs, but not images: some images are shared between training and testing data. 

\begin{figure}[t]
    \begin{center}
      \includegraphics[width=1.0\linewidth]{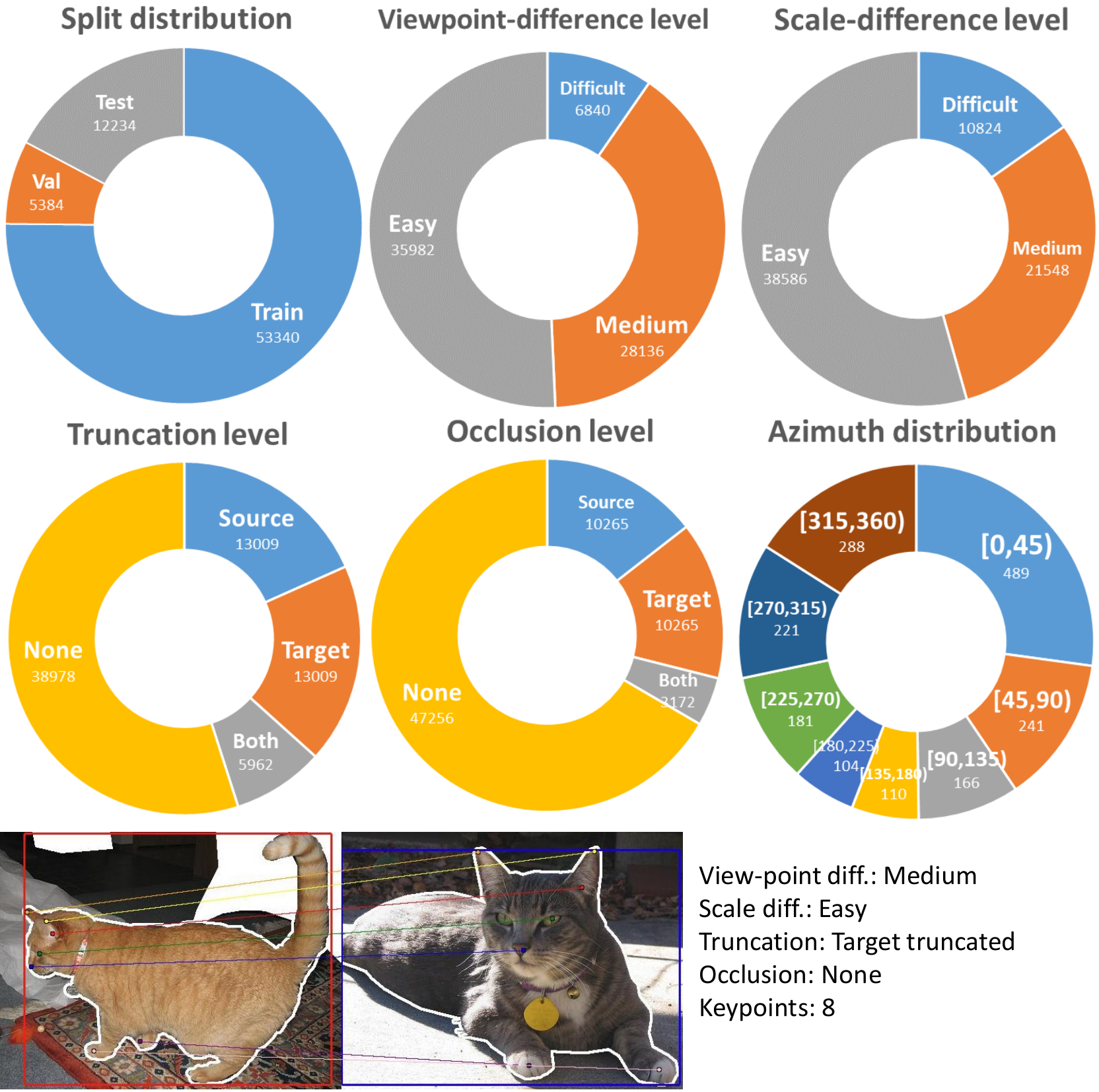}
    \end{center}
    \vspace{-6mm}
    \caption{SPair-71k data statistics and an example pair with its annotations. Best viewed in electronic form.}
    \label{fig:dataset_summary_example}
    \vspace{-5mm}
\end{figure}

To resolve these issues, we introduce a new dataset, {\em SPair-71k}, consisting of total 70,958 pairs of images from PASCAL 3D+~\cite{Xiang2014BeyondPA} and PASCAL VOC 2012~\cite{everingham2015pascal}\footnote{We do not include `dining table' and `sofa' classes because they appear as background in most images and their semantic keypoints are too ambiguous to localize.}. The dataset is significantly larger with rich annotations and clearly organized for learning.   
%No image has overlap on this dataset train/val/test split. 
In particular, several types of useful annotations are available: keypoints of semantic parts, object segmentation masks, bounding boxes, view-point, scale, truncation, and occlusion differences for image pairs, etc. %~\cite{everingham2010pascal,Xiang2014BeyondPA}. We've also imported the azimuth-level annotation from ~\cite{Xiang2014BeyondPA} and manually annotated keypoints.
%Using image-level annotation, we added four challenges as pair-level annotations.
Figure~\ref{fig:dataset_summary_example} presents the statistics of SPair-71k in pie chart forms and shows a sample image pair with its annotations. For details on our dataset, we refer the readers to the website: {\tt\small  \url{http://cvlab.postech.ac.kr/research/SPair-71k/}}. 
\vspace{-3mm}

\begin{center}
    \begin{table*}
        \begin{center}
            \scalebox{0.85}{
                \begin{tabular}{l|c|ccc|ccc|cc}
                \hline
                \multirow{2}{*}{Methods} &\multirow{2}{*}{Supervision}  & \multicolumn{3}{c|}{PF-PASCAL (PCK@$\alpha_{\text{img}}$)} & \multicolumn{3}{c|}{PF-WILLOW (PCK@$\alpha_{\text{bbox}}$)} & \multicolumn{2}{c}{Caltech-101} \\   
                 & & 0.05 & 0.1 & 0.15 & 0.05 & 0.1 & 0.15 & LT-ACC & IoU \\ 
                \hline
                \hline
                 Identity mapping                   & \multirow{2}{*}{-} & 12.7 & 37.0 & 60.8 & 12.2 & 27.0 & 41.7 & 0.77 & 0.44 \\
                %\cline{2-10}%
                % \hline
                PF$_\textrm{HOG}$~\cite{ham2016proposal} &   & 31.4 & 62.5 & 79.5 & 28.4 & 56.8 & 68.2 & 0.78 & 0.50 \\  
                \hline%\cline{1-1}            
                 CNNGeo$_\textrm{res101}$~\cite{Rocco17}      & \multirow{2}{*}{\makecell{synthetic warp \\ \small{(self-supervised)}}} & 41.0 & 69.5 & 80.4 & 36.9 & 69.2 & 77.8 & 0.79 & 0.56 \\  
                 A2Net$_\textrm{res101}$~\cite{paul2018attentive}    &  & 42.8 & 70.8 & 83.3 & 36.3 & 68.8 & 84.4 & 0.80 & 0.57 \\  
                \hline%\cline{1-1}
                 DCTM$_\textrm{CAT-FCSS}$~\cite{kim2017fcss}   & \multirow{4}{*}{\makecell{image labels \\ \small{(weakly-supervised)}}}  & 34.2 & 69.6 & 80.2 & 38.1 & 61.0 & 72.1 & 0.83 & 0.52 \\  
                Weakalign$_\textrm{res101}$~\cite{Rocco18}        &   &  49.0 & 74.8 & 84.0 & 37.0 & 70.2 & 79.9 & 0.85 & \underline{0.63} \\  
                NC-Net$_\textrm{res101}$~\cite{rocco2018neighbourhood} & & 54.3 & 78.9 & 86.0 & 33.8 & 67.0 & 83.7 & 0.85 & 0.60  \\  
                RTNs$_\textrm{res101}$~\cite{NIPS2018_7851} & & 55.2 & 75.9 & 85.2 & 41.3 & 71.9 & \underline{86.2} & - & - \\  
                \hline% \cline{1-1}
                UCN$_\textrm{GoogLeNet}$~\cite{choy2016universal}  & \multirow{3}{*}{keypoints} & 29.9 & 55.6 & 74.0 & 24.1 & 54.0 & 66.5 & - & - \\  
                SCNet$_\textrm{vgg16}$~\cite{han2017scnet} &  & 36.2 & 72.2 & 82.0 & 38.6 & 70.4 & 85.3 & 0.79 & 0.51 \\ NN-Cyc$_\textrm{res101}$~\cite{laskar2019semantic}  & & 55.1 & \underline{85.7} & \underline{94.7} & 40.5 & \underline{72.5} & \underline{86.9} & 0.86 & 0.62 \\  
                \hline   
                HPF$_{\textrm{res50}}$ (ours)  & \multirow{3}{*}{\makecell{keypoints \\ \small{(validation only)}}}  & \underline{60.5} & 83.4 & 92.1 & \underline{46.5} & 72.4 & 84.7 & \textbf{0.88} & \textbf{0.64} \\
                HPF$_{\textrm{res101}}$ (ours)  &  & \underline{60.1} & \underline{84.8} & \underline{92.7} & \underline{45.9} & \underline{74.4} & 85.6 & \underline{0.87} & \underline{0.63} \\
                HPF$_{\textrm{res101-FCN}}$ (ours) & & \textbf{63.5} & \textbf{88.3} & \textbf{95.4} & \textbf{48.6} & \textbf{76.3} & \textbf{88.2} & \underline{0.87} & \underline{0.63} \\ 
                % \cline{1-1}\cline{3-10}
                \hline
                \hline
                HPF$_{\textrm{res101}}$ ($k$=1)  & \multirow{3}{*}{\makecell{keypoints \\ \small{(validation only,} \\\small{small set)}}} & 59.4$_{\pm0.89}$ & 83.9$_{\pm1.14}$ & 92.2$_{\pm0.99}$ & 44.5$_{\pm0.90}$ & 72.5$_{\pm1.22}$ & 84.8$_{\pm0.93}$ & 0.87 & 0.63 \\
                HPF$_{\textrm{res101}}$ ($k$=2)   &  & 58.3$_{\pm1.33}$ & 84.5$_{\pm0.77}$ & 92.9$_{\pm0.41}$ & 44.7$_{\pm0.92}$ & 73.1$_{\pm1.05}$ & 85.4$_{\pm0.84}$ & 0.87 & 0.63 \\
                HPF$_{\textrm{res101}}$ ($k$=3)  & & 59.4$_{\pm1.16}$ & 84.5$_{\pm0.27}$ & 92.7$_{\pm0.35}$ & 45.1$_{\pm0.55}$ & 73.4$_{\pm0.52}$ & 85.4$_{\pm0.48}$ & 0.87 & 0.63 \\
                \hline
                HPF$_{\textrm{res101}}$ (random) & - & 44.5$_{\pm11.11}$ & 74.7$_{\pm6.46}$ & 87.3$_{\pm3.13}$ & 32.8$_{\pm8.12}$ & 62.4$_{\pm6.67}$ & 78.2$_{\pm4.20}$ & 0.85 & 0.55 \\
                \hline
                \end{tabular}
            }
            \vspace{-1.5mm}
            \caption{\label{tab:stdBenchmarkTable}Results on standard benchmarks of semantic correspondences. Subscripts of the method names indicate  backbone networks used.  The second column denotes supervisory information used for training or tuning.
            Numbers in bold indicate the best performance and underlined ones are the second and third best. Results of~\cite{ham2016proposal, han2017scnet, kim2017fcss, Rocco17, Rocco18} are borrowed from~\cite{NIPS2018_7851}.
            }
            \vspace{-3.5mm}
        \end{center}
    \end{table*}
\end{center}

\section{Experimental Evaluation}
%We conducted extensive experiments on standard benchmarks, PF-PASCAL, PF-WILLOW, Caltech-101 as well as our new benchmark dataset, SPair-71k. 
In this section we compare the proposed method with recent state-of-the-art methods and discuss the results.

\subsection{Implementation details}
We use two CNNs as main backbone networks for hyperpixel features, ResNet-50 and ResNet-101~\cite{he2016deep} pre-trained on ImageNet~\cite{deng2009imagenet}. All convolutional layers of the networks are used as candidate feature layers for hyperpixels. We extract the features at the end of each layer before a ReLU activation. The optimal set of hyperpixel layers, $(l_0, ..., l_{K-1})$,  
%Tand noise suppressing constant $d$ of the appearance similarity term, 
is determined by Algorithm~\ref{alg:layersearch} run with a validation split of a target dataset. For this beam search, we set the beam size 4 and the maximum number of layers allowed 8.
%with varying values of $d$ such that $1 \leq d \leq 5$. 
For the exponent value for hyperpixel similarity, we fix $d=3$ based on search using PF-PASCAL validation split. 

\subsection{Evaluation metric}\label{sec:PCK}
For evaluation on PF-WILLOW, PF-PASCAL, and SPair-71k, we use a common evaluation metric of percentage of correct keypoints (PCK), which counts the average number of correctly predicted keypoints given a tolerance threshold. Given predicted keypoint $\mathbf{k}_{\mathrm{pr}}$ and ground-truth keypoint $\mathbf{k}_{\mathrm{gt}}$, the prediction is considered correct if Euclidean distance between them is smaller than a given threshold. The correctness $c$ of each keypoint can be expressed as 
\begin{align}
    c = \begin{cases}
    1 \ \ \ \ \text{if} \quad d(\mathbf{k}_{\mathrm{pr}}, \mathbf{k}_{\mathrm{gt}}) \leq \alpha_{\tau} \cdot \max{(w_{\tau}, h_{\tau})} \\
    0 \ \ \ \ \text{otherwise},
    \end{cases}
\end{align}
where $w_{\tau}$ and $h_{\tau}$ are the width and height of either an entire image or object bounding box, $\tau \in \{\text{img}, \text{bbox}\}$, and $\alpha_{\tau}$ is a tolerance factor (in most cases, $\alpha = 0.1$). Note that PCK with $\alpha_{\mathrm{bbox}}$ is a more stringent metric than one with $\alpha_{\mathrm{img}}$. The final PCK of a benchmark is evaluated by averaging PCKs of all input image pairs. Following recent papers~\cite{han2017scnet, Rocco17, Rocco18, rocco2018neighbourhood}, we evaluate PF-WILLOW with $\alpha_{\mathrm{bbox}}$ and PF-PASCAL with $\alpha_{\mathrm{img}}$ using the same dataset split as in~\cite{rocco2018neighbourhood}. For SPair-71k, we use $\alpha_{\mathrm{bbox}}$, which is more stringent. %, which we think is more reaonable. %This is a more appropriate method for semantic correspondence that focuses on the semantic part of the object.
% \jmlee{We also used PCK to evaluate the SPair-71k. However, at the time of PCK evaluation, evaluation can be done by applying two conditions, threshold and average criteria. First, we threshold with bounding box size. This is a more appropriate method for semantic correspondence that focuses on the semantic part of the object. Second, we average pck by pair and then evaluate the value. For the consistency of evaluation, we did not compute the average PCK with the numbers of correct keypoints / the numbers of total keypoints.}

\begin{figure}[t]
    \begin{center}
    \includegraphics[width=1.0\linewidth]{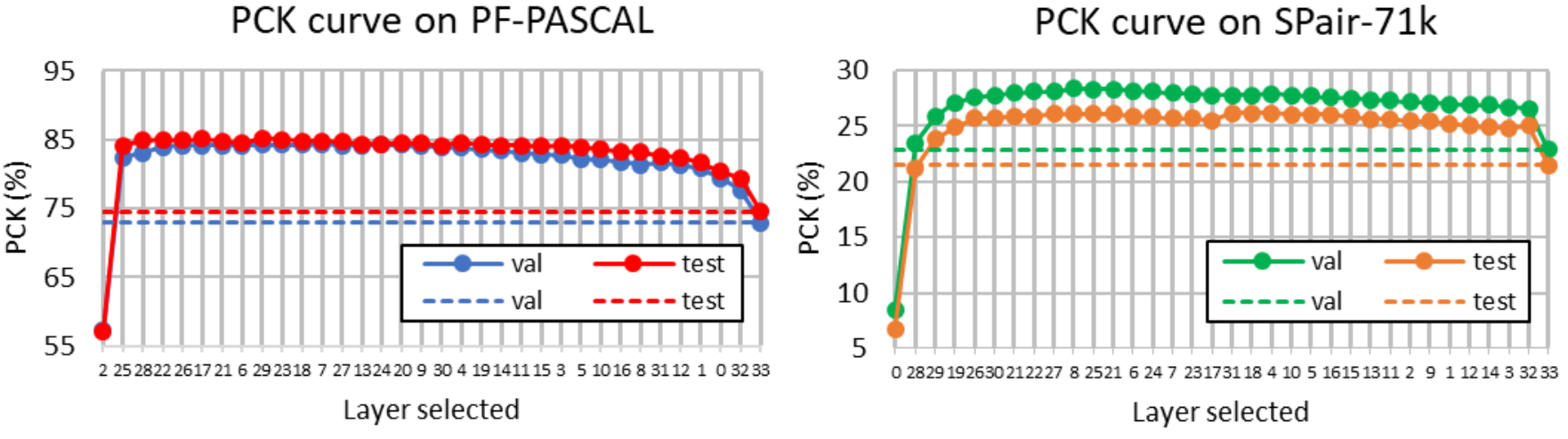}
    \end{center}
     \vspace{-5.5mm}
    \caption{Hyperpixel layer search with ResNet-101 backbone on PF-PASCAL and SPair-71k datasets. Hyperpixel layers are in the order of selection during beam search. Dashed lines indicate PCKs when all layers of a CNN are used for hyperpixels. Best viewed in electronic form.}
    \vspace{-5.0mm}
    \label{fig:hyperpixelAblation}
\end{figure}

\begin{center}
    \begin{table*}
        \begin{center}
            \scalebox{0.65}{
            \begin{tabular}{c|c|cccccccccccccccccc|c}
            \hline
             \multicolumn{2}{c|}{Methods} & aero & bike & bird & boat & bottle & bus & car & cat & chair & cow & dog & horse & moto & person & plant & sheep & train & tv & all\\
            % \hline
            \hline
            % \multicolumn{2}{c|}{Identity Mapping} &  &  &  &  &  &  &  &  &  &  &  &  &  &  &  &  &  &  &  \\
            \hline
            \multirow{4}{*}{\shortstack[1]{Transferred \\ \\ models}}
            & CNNGeo$_\textrm{res101}$~\cite{Rocco17} & 21.3 & 15.1 & 34.6 & 12.8 & 31.2 & 26.3 & 24.0 & 30.6 & 11.6 & 24.3 & 20.4 & 12.2 & 19.7 & 15.6 & 14.3 & 9.6 & 28.5 & 28.8 & 18.1  \\
            & A2Net$_\textrm{res101}$~\cite{paul2018attentive} &  20.8 & 17.1 & 37.4 & 13.9 & 33.6 & \underline{29.4} & \underline{26.5} & 34.9 & 12.0 & 26.5 & 22.5 & 13.3 & 21.3 & 20.0 & 16.9 & 11.5 & 28.9 & 31.6 & 20.1 \\
            & WeakAlign$_\textrm{res101}$~\cite{Rocco18} & 23.4 & 17.0 & 41.6 & 14.6 & 37.6 & \underline{28.1} & \underline{26.6} & 32.6 & 12.6 & 27.9 & 23.0 & 13.6 & 21.3 & 22.2 & 17.9 & 10.9 & \underline{31.5} & 34.8 & 21.1 \\
            & NC-Net$_\textrm{res101}$~\cite{rocco2018neighbourhood} & \underline{24.0} & 16.0 & \underline{45.0} & 13.7 & 35.7 & 25.9 & 19.0 & \underline{50.4} & \underline{14.3} & \underline{32.6} & \underline{27.4} & \underline{19.2} & \underline{21.7} & 20.3 & 20.4 & \underline{13.6} & \textbf{33.6} & \textbf{40.4} & \underline{26.4} \\
            % HPF$_\mathrm{res101}$ (ours) & \textbf{26.9} & \textbf{20.9} & \textbf{50.1} & \textbf{17.2} & \textbf{40.1} & 23.7 & 20.3 & 50.4 & \textbf{19.3} & 32.8 & \textbf{31.8} & 21.2 & 24.2 & \textbf{26.0} & \textbf{23.4} & \textbf{17.7} & 31.4 & 37.3 & 28.8  \\
            \hline
            \multirow{4}{*}{\shortstack[1]{SPair-71k \\ \\ trained \\ \\  models}}
            & CNNGeo$_\textrm{res101}$~\cite{Rocco17} &  23.4 & 16.7 & 40.2 & 14.3 & 36.4 & 27.7 & 26.0 & 32.7 & 12.7 & 27.4 & 22.8 & 13.7 & 20.9 & 21.0 & 17.5 & 10.2 & 30.8 & 34.1 & 20.6  \\
            & A2Net$_\textrm{res101}$~\cite{paul2018attentive} & 22.6 & \underline{18.5} & 42.0 & \textbf{16.4} & \underline{37.9} & \textbf{30.8} & \underline{26.5} & 35.6 & 13.3 & 29.6 & 24.3 & 16.0 & 21.6 & \underline{22.8} & \underline{20.5} & 13.5 & 31.4 & \underline{36.5} & 22.3 \\
            & WeakAlign$_\textrm{res101}$~\cite{Rocco18} &  22.2 & 17.6 & 41.9 & \underline{15.1} & \textbf{38.1} & 27.4 & \textbf{27.2} & 31.8 & 12.8 & 26.8 & 22.6 & 14.2 & 20.0 & 22.2 & 17.9 & 10.4 & \underline{32.2} & 35.1 & 20.9 \\
            & NC-Net$_\textrm{res101}$~\cite{rocco2018neighbourhood} & 17.9 & 12.2 & 32.1 & 11.7 & 29.0 & 19.9 & 16.1 & 39.2 & 9.9 & 23.9 & 18.8 & 15.7 & 17.4 & 15.9 & 14.8 & 9.6 & 24.2 & 31.1 & 20.1   \\
            % \cline{2-21}
            \hline
            % \multicolumn{2}{c|}{ HPF$_\mathrm{res50}$ (ours)}  & \textbf{25.4} & \textbf{18.9} & \underline{47.5} & 14.6 & 37.4 & 22.7 & 17.6 & \underline{50.8} & \underline{16.6} & 32.2 & \underline{31.1} & \underline{19.2} & \underline{23.0} & \underline{24.0} & \textbf{23.7} & \underline{13.9} & 30.1 & \underline{37.4} & \underline{27.2} \\
            % \multicolumn{2}{c|}{HPF$_\mathrm{res101}$ (ours)}  & \underline{25.3} & 18.4 & \textbf{52.5} & \underline{15.9} & \textbf{38.3} & 22.1 & 18.6 & \textbf{53.1} & \textbf{18.0} & \textbf{33.4} & \textbf{32.9} & \textbf{20.5} & \textbf{24.1} & \textbf{27.3} & \underline{21.2} & \textbf{16.2} & 31.1 & 36.1 & \textbf{28.2} \\
            \multicolumn{2}{c|}{ HPF$_\mathrm{res50}$ (ours)}  & \textbf{25.3} & \underline{18.5} & \underline{47.6} & 14.6 & 37.0 & 22.9 & 18.3 & \underline{51.1} & \underline{16.7} & \underline{31.5} & \underline{30.8} & \underline{19.1} & \underline{23.7} & \underline{23.8} & \textbf{23.5} & \underline{14.4} & 30.8 & \underline{37.2} & \underline{27.2} \\
            \multicolumn{2}{c|}{HPF$_\mathrm{res101}$ (ours)}  & \underline{25.2} & \textbf{18.9} & \textbf{52.1} & \underline{15.7} & \underline{38.0} & 22.8 & 19.1 & \textbf{52.9} & \textbf{17.9} & \textbf{33.0} & \textbf{32.8} & \textbf{20.6} & \textbf{24.4} & \textbf{27.9} & \underline{21.1} & \textbf{15.9} & \underline{31.5} & 35.6 & \textbf{28.2} \\
            \hline
            \end{tabular}}
            \vspace{-1.5mm}
        \caption{\label{tab:hpftable}Per-class PCK ($\alpha_{\text{bbox}}=0.1$) results on SPair-71k dataset. For transferred model, the original models trained on PASCAL-VOC~\cite{Rocco17, paul2018attentive} and PF-PASCAL~\cite{Rocco18, rocco2018neighbourhood}, which are provided by the authors, are used for evaluation.
        %HPF$_\mathrm{res50}$ and HPF$_\mathrm{res101}$ are tuned with SPair-71k validation set.
        Note that, for SPair-71k trained models, the transferred models are further finetuned on SPair-71k dataset by ourselves with our best efforts. Numbers in bold indicate the best performance and underlined ones are the second and third best.}
        % CNNGeo~\cite{Rocco17} and A2Net~\cite{paul2018attentive} are trained with InlierCount loss~\cite{Rocco18}.} 
        \vspace{-4.5mm}
        % since SPair-71k dataset does not have geometric transformation parameter supervision
        % \cite{Rocco18, rocco2018neighbourhood} have lower scores after training because their models are already trained on PF-PASCAL training set. 
        % the Result of weakalgin is lower than GeoCNN, because weakalign model overfit to the PF-PASCAL training set.
        % Note that weakalign [37] model has lower score tooverfit PF-PASCAL training dataset.
        \end{center}
    \end{table*}
\end{center} 

\begin{table}[t]
    \begin{center}
        \scalebox{0.9}{
            \begin{tabular}{l|c|cc}
                \hline
                Approach & Model & PCK & Time (\em ms) \\
                \hline\hline
                \multirow{4}{*}{\shortstack[1]{Image \\ alignment}} 
                 & CNNGeo$_\textrm{res101}$~\cite{Rocco17}                & 69.5 & \underline{40} \\
                 & WeakAlign$_\textrm{res101}$~\cite{Rocco18}             & 74.8 & 41 \\
                 & A2Net$_\textrm{res101}$~\cite{paul2018attentive}       & 70.8 & 53 \\
                 & RTNs$_\textrm{res101}$~\cite{NIPS2018_7851}            & 75.9 & 376 \\ 
                \hline
                \multirow{7}{*}{\shortstack[1]{Local \\ \\ region \\ \\ matching}} 
                 & SCNet$_\textrm{vgg16}$~\cite{han2017scnet}            & 72.2 & $>$ 1000 \\
                 & PF$_\textrm{HOG}$~\cite{ham2016proposal}            & 62.5 & $>$ 1000 \\
                 & NC-Net$_\textrm{res101}$~\cite{rocco2018neighbourhood} & 78.9 & 261  \\\cline{2-4}
                 & HPF$_\textrm{res101}$ w/ all layers & 74.5 & 324 \\
                 & HPF$_\textrm{res50}$ w/ all layers & 70.1 & 130 \\
                 & HPF$_\textrm{res101}$    & \textbf{84.8} & 63 \\
                 & HPF$_\textrm{res50}$     & \underline{83.4} & \underline{34} \\
                 & HPF$_\textrm{res50}*$    & \underline{81.1} & \textbf{19} \\
                \hline
            \end{tabular}
        }
    \end{center}
    \vspace{-4mm}
    \caption{\label{tab:ablationSpeed}Inference time comparison on PF-PASCAL benchmark. Hyperpixel layers of HPF$_\textrm{res50}*$ are (4,7,11,12,13).}
    % while models with star sign ($^{\star}$) uses all layers of a CNN.}
    \vspace{-2mm}
\end{table}

\subsection{Results and analysis}

\smallbreak
\noindent \textbf{Hyperpixel layers.} 
For PF-PASCAL, the hyperpixel layer results are $(2,7,11,12,13)$ with ResNet-50 and $(2,17,21,22,25,26,28)$ with ResNet-101. For SPair-71k, the results are $(0,9,10,11,12,13)$ with ResNet-50 and $(0,8,20,21,26,28,29,30)$ with ResNet-101. %To verify this layer setting is optimal, we performed a grid search over all the layers and the difference was trivial. 
In order to analyze the effect of each intermediate feature $(\mathbf{f}^{l_0},..., \mathbf{f}^{l_{K-1}})$ on hyperpixel, we have measured PCK of our model on both PF-PASCAL and SPair-71k in the order of the layer selection during beam search as shown in Figure~\ref{fig:hyperpixelAblation}. The dashed lines represent PCKs using all layers. Interestingly, in both cases, adding the second layer significantly boosts the performance of PCK, and only a few more layers are sufficient to achieve a comparable performance with the best one. 
After reaching an optimized set of layers, adding more damages the performance. 
This result demonstrates the effectiveness of hyperpixels compared to conventional hypercolumn features. 
The result also implies that features resolving local-ambiguity lie in between particular layers, \eg, between layer 20 and 30 in our case. 
%  shows This convergence and a huge performance drop with all layers imply that simply adding more semantics to hyperpixels entails only noisy information. 

\smallbreak
\noindent \textbf{Benchmark comparisons.} 
Table~\ref{tab:stdBenchmarkTable} summarizes comparison to recent methods on three standard benchmarks: PF-PASCAL, PF-WILLOW, and Caltech-101. In this experiment, the hyperpixels tuned using the validation split of PF-PASCAL are evaluated on the test split of PF-PASCAL, and futher evaluated on PF-WILLOW and Caltech-101 for checking transferability as done in~\cite{han2017scnet, jeon2018parn, NIPS2018_7851, Rocco17, Rocco18, paul2018attentive}. 
The results clearly show that the proposed method sets new state-of-the-art results on all the three benchmarks, proving the effectiveness of our approach. Note that all recent neural methods for semantic correspondence rely 
on ImageNet-pretrained features, and thus their performance depends on the backbone networks (indicated by subscripts).
As expected, our method using the stronger backbone of ResNet-101 improves the performance compared to using ResNet-50. 
Furthermore, using the backbone of FCN~\cite{lin2017feature} pretrained with PASCAL VOC 2012~\cite{everingham2015pascal}, that is a superset of our target dataset~\cite{ham2016proposal}, significantly boosts performance.  
This shows that our method is flexible in using backbone networks and can further improve by adopting a better one.

\smallbreak
\noindent \textbf{Degree of supervision.} 
Different methods in our comparison require different degrees of supervision in training as indicated in the second column of Table~\ref{tab:stdBenchmarkTable}. 
The only supervised part of our method is layer selection using a validation set, which can be very small as revealed by small-set experiments, and does not require additional learning: 
% Instead of using all the 308 pairs of the original validation split of PF-PASCAL, we have used $k$ random pairs per class, for a total of $20k$ validation pairs, to select optimal layers. 
Instead of using all the 308 pairs of the original validation split of PF-PASCAL, 
the layer search algorithm is performed on $k$ random pairs per class, for a total of $20k$ validation pairs. 
The average performances over 10 trials are shown along with their standard deviations in the set of rows with $k=1,2,3$ at the bottom of Table~\ref{tab:stdBenchmarkTable}.
Using as little as one sample per class (20 image pairs total) as supervisory signal gives results comparable as using all 308 pairs, outperforming the previous state of the art. 
Given the cost of data collection and the total amount of user-provided information in weakly-supervised methods, we thus believe that our algorithm with small $k$ values (e.g., $k=1$) is more cost effective and practical.
%\jmin{To validate the practicability and effectiveness of our approach, we perform {\em small-set} experiments by running Alg.~\ref{alg:layersearch} on a significantly fewer number of validation pairs;
%only $k$ random pairs per class are used to select optimal layers instead of all the 308 pairs.
%As shown in Table~\ref{tab:stdBenchmarkTable}, using as little as one sample per class (20 image pairs total) as supervisory signal gives results, averaged over 10 trials, comparable as using all 308 pairs, outperforming the previous state of the art\footnote{Note that the level of supervision in~\cite{laskar2019semantic} is {\em much} higher with 4300 image pairs and keypoint annotations.}.}

\smallbreak
\noindent \textbf{Effect of layer search.} To check the effect of layer search, we take random combinations of 8 layers (the same number chosen by our layer search) as a baseline. 
% The result, averaged over 10 trials, is in the last row of Table~\ref{tab:stdBenchmarkTable}. 
The average results over 10 trials are shown with their standard deviations in the last row of Table~\ref{tab:stdBenchmarkTable}.
%and is much worse than using our layer selection method.
%Following R2's, we evaluated the baseline of random selection by replacing our layer search with a random combination of 7 layers (the same \# of layers in the result of our layer search). The average result over 10 trials is shown in the last row of Table~\ref{tab:table_1}.
Their much worse performance shows that our layer search is crucial. 

\begin{table}[t]
    \begin{center}
        \scalebox{0.95}{
        \begin{tabular}{l|c|c}
            \hline
            \multirow{2}{*}{Matching module} & PF-PASCAL & PF-WILLOW \\
             & $\alpha_{\textrm{img}}=0.1 $ & $\alpha_{\textrm{bbox}}=0.1$ \\
            \hline
            \hline
            NN w/ $(d=1)$  & 69.0 & 60.9 \\
            RHM w/ $(d=1)$ & 81.4 & 68.6 \\
            RHM w/ $(d=2)$ & 84.4 & 73.3 \\
            \rowcolor{mygray} RHM w/ $(d=3)$* & \textbf{84.8} & \textbf{74.4} \\
            RHM w/ $(d=4)$ & \underline{84.8} & \underline{74.1} \\
            RHM w/ $(d=5)$ & \underline{84.5} & \underline{73.9} \\
            \hline
        \end{tabular}}
        \vspace{-1.5mm}
    \caption{\label{tab:ablationModule}Ablation studies on RHM with ResNet-101.}
    \vspace{-8.0mm}
    \end{center}
\end{table}

\begin{figure*}[htbp]
    \centering
    \begin{subfigure}{0.144\textwidth}
        \includegraphics[width=1.0\linewidth, keepaspectratio]{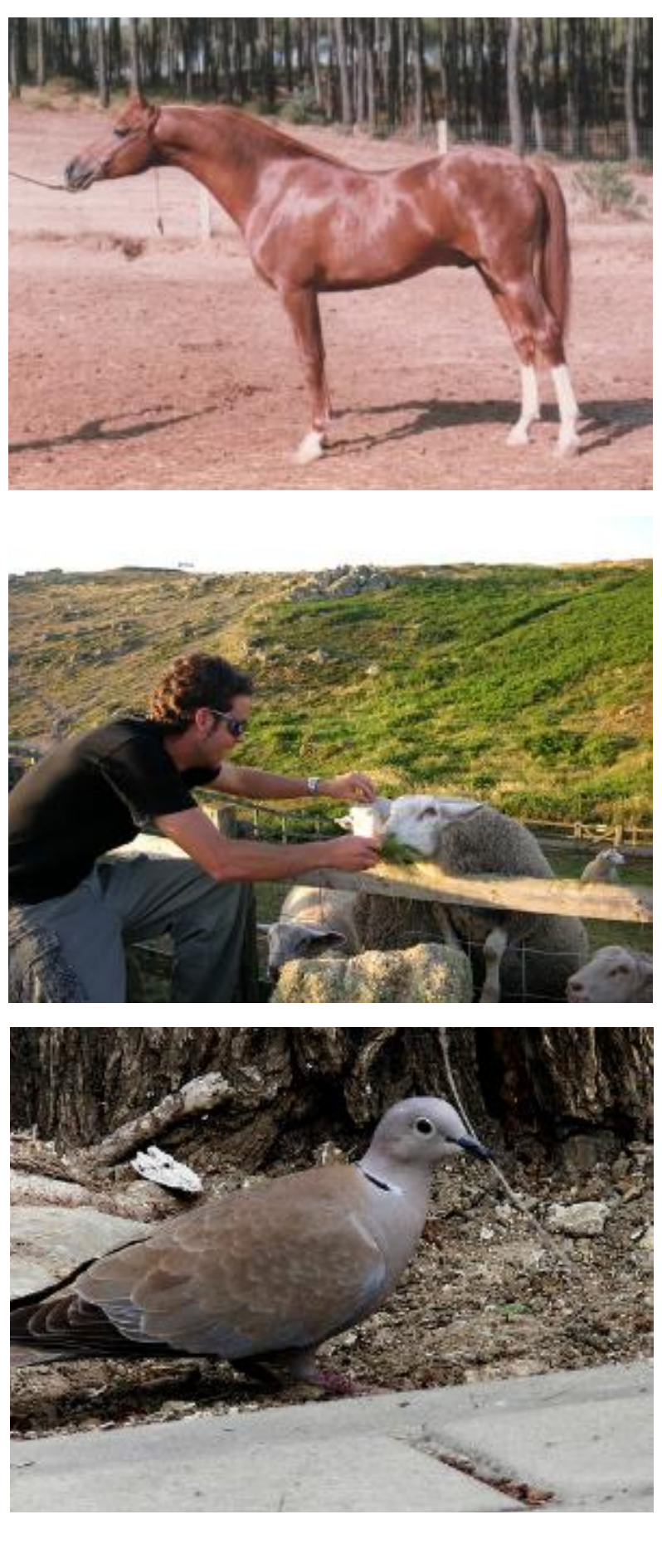} 
        \caption{Source image}
    \end{subfigure}
    \hspace{-2.0mm}
    \begin{subfigure}{0.135\textwidth}
        \includegraphics[width=1.0\linewidth, keepaspectratio]{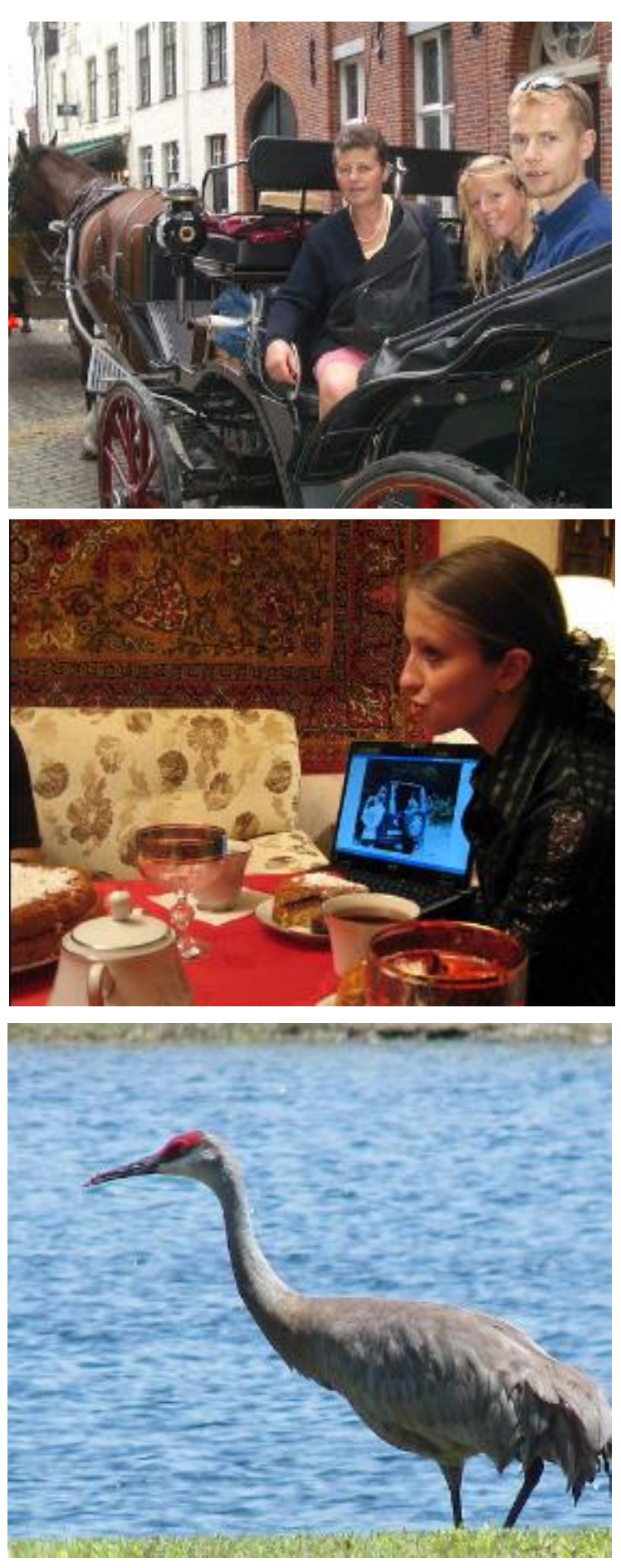}
        \caption{Target image}
    \end{subfigure}
    \hspace{-2.0mm}
    \begin{subfigure}{0.144\textwidth}
        \includegraphics[width=1.0\linewidth, keepaspectratio]{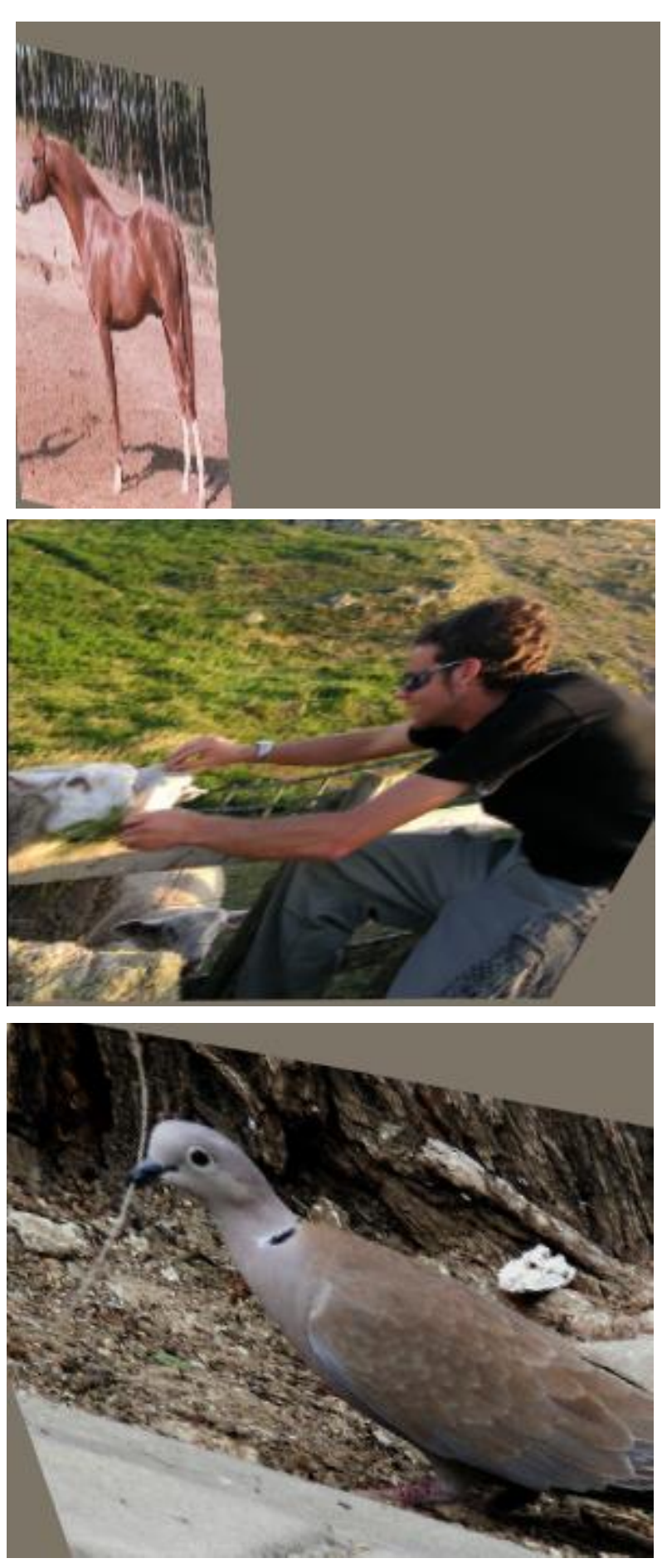}
        \caption{HPF (ours)}
    \end{subfigure}
    \hspace{-2.0mm}
    \begin{subfigure}{0.144\textwidth}
        \includegraphics[width=1.0\linewidth, keepaspectratio]{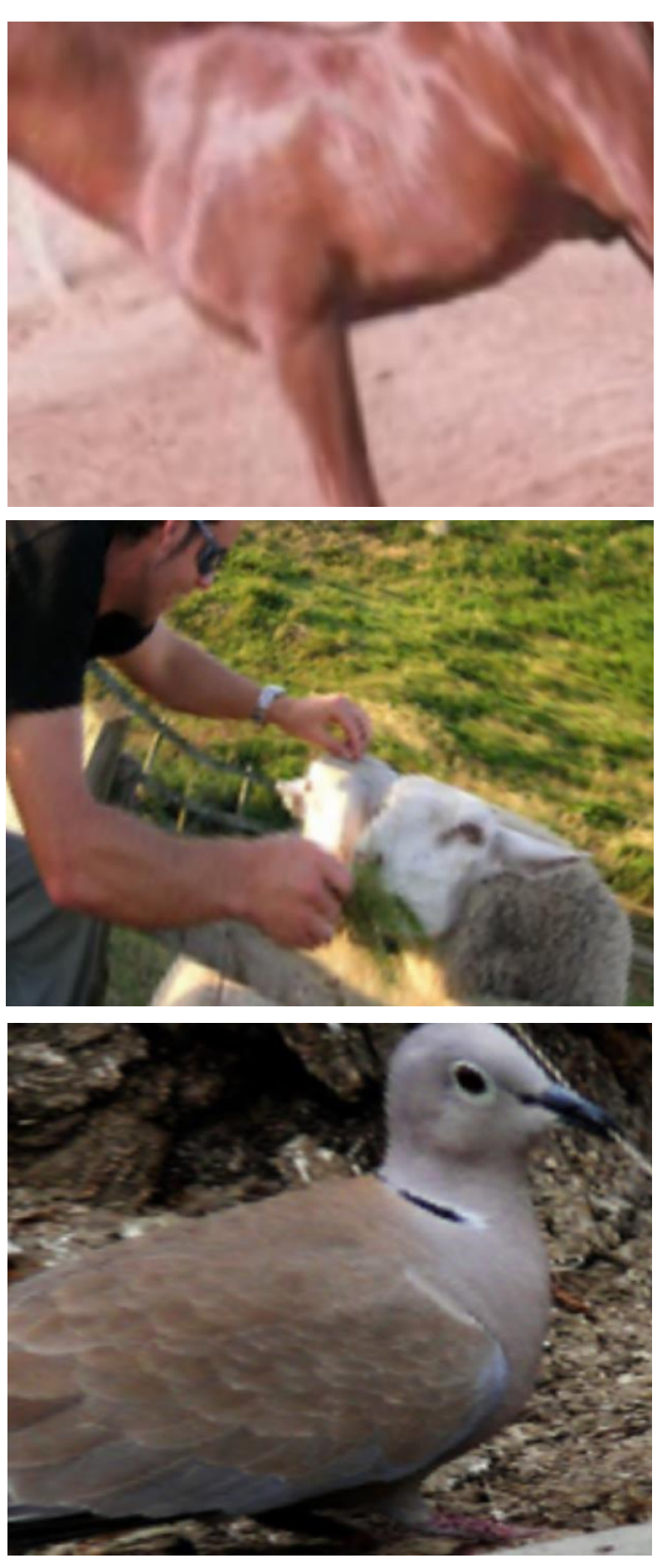}
        \caption{CNNGeo~\cite{Rocco17}}
    \end{subfigure}
    \hspace{-2.0mm}
    \begin{subfigure}{0.144\textwidth}
        \includegraphics[width=1.0\linewidth, keepaspectratio]{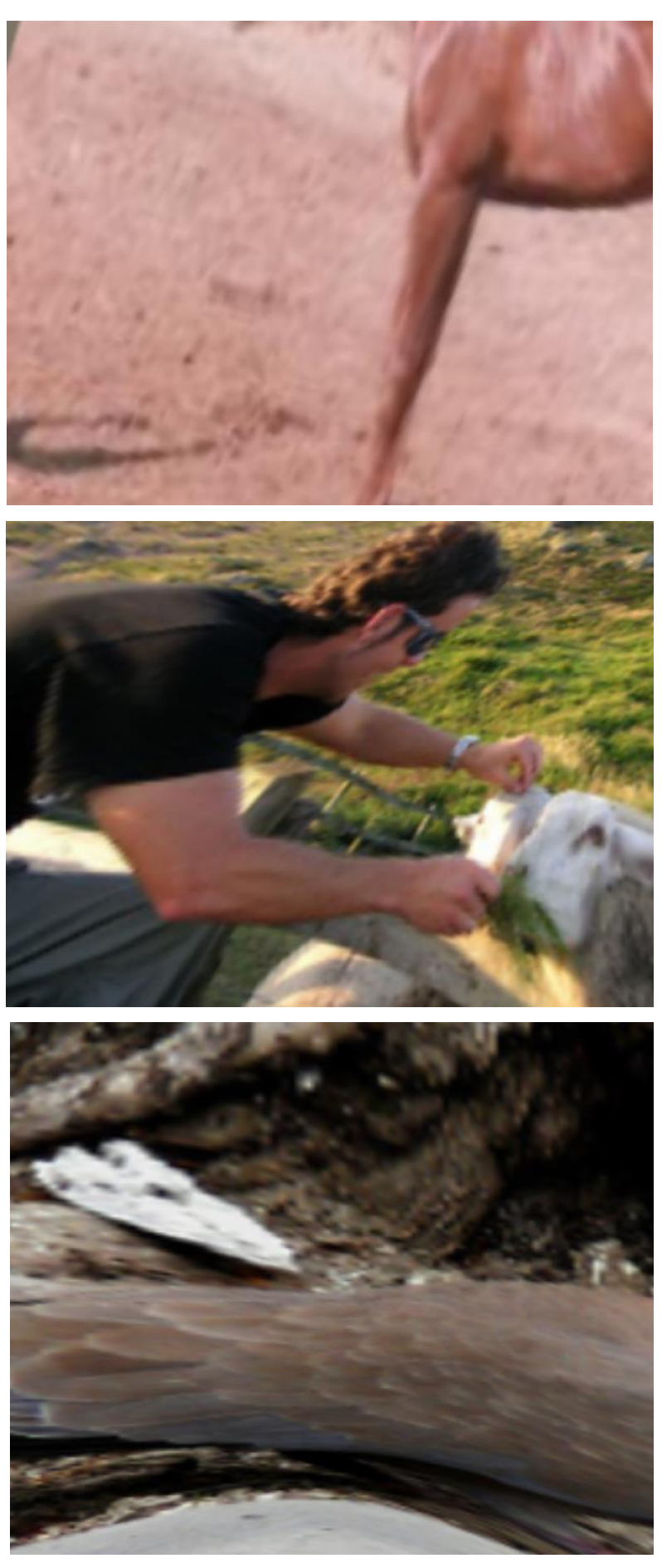}
        \caption{A2Net~\cite{paul2018attentive}}
    \end{subfigure}
    \hspace{-2.0mm}
    \begin{subfigure}{0.144\textwidth}
        \includegraphics[width=1.0\linewidth, keepaspectratio]{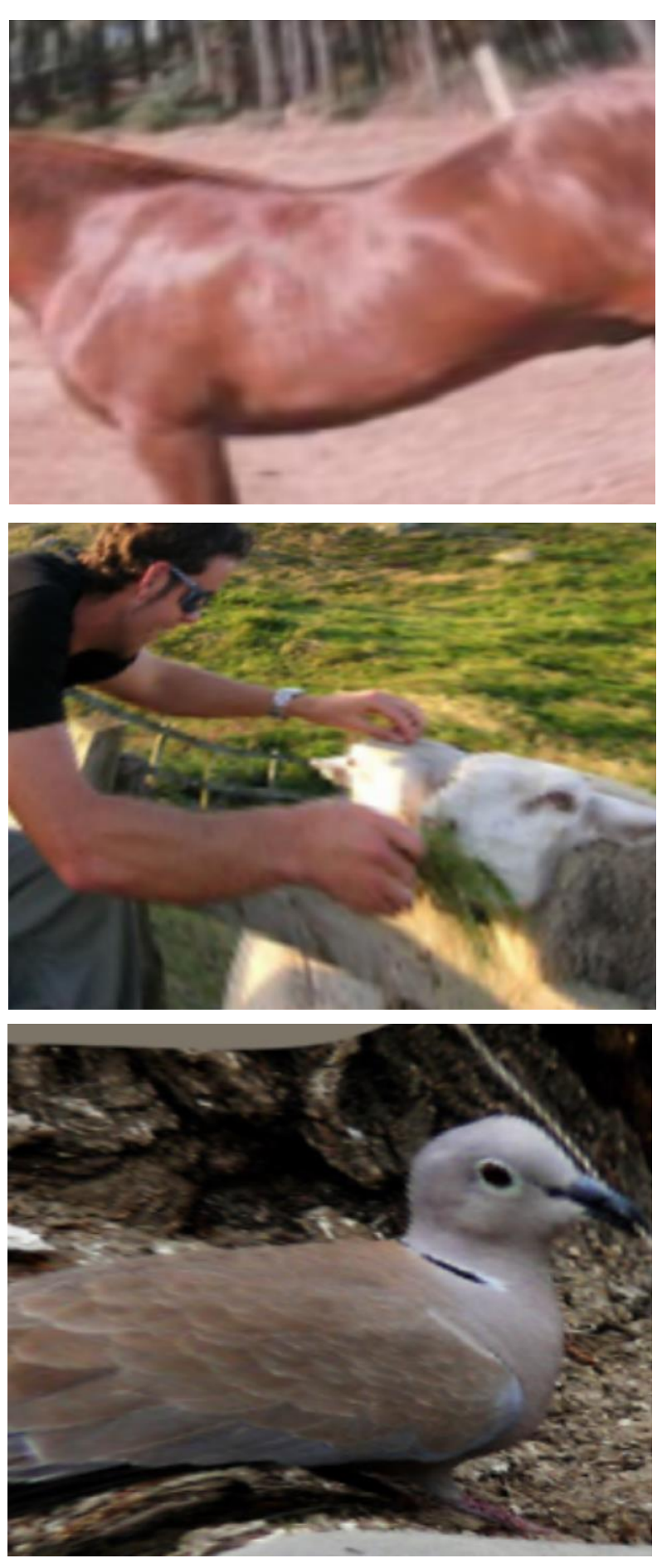}
        \caption{WeakAlign~\cite{Rocco18}}
    \end{subfigure}
    \hspace{-2.0mm}
    \begin{subfigure}{0.144\textwidth}
        \includegraphics[width=1.0\linewidth, keepaspectratio]{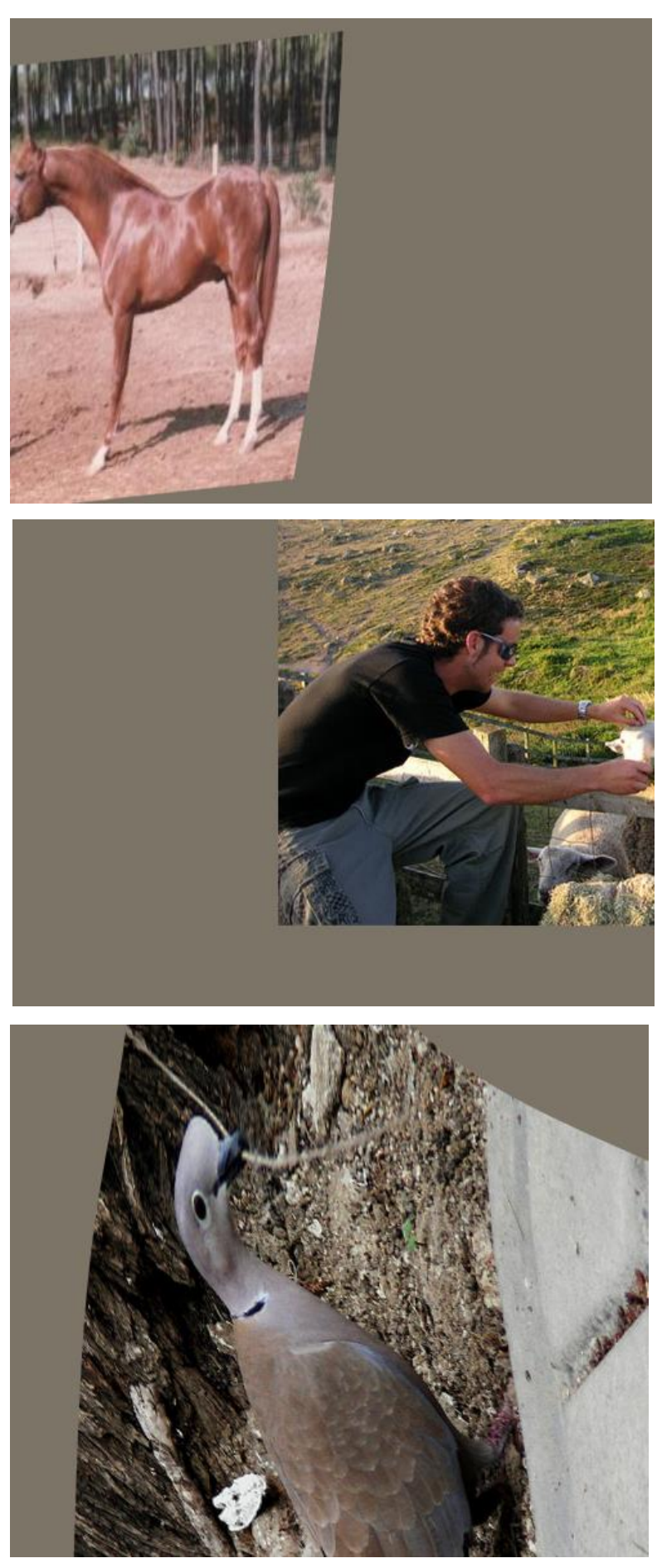}
        \caption{NC-Net~\cite{rocco2018neighbourhood}}
    \end{subfigure}
    \hspace{-2.0mm}
    \vspace{-3.0mm}
    \caption{Qualitative results on SPair-71k. The source images are transformed to target images using correspondences.}
    \label{fig:figure4}
    \vspace{-3.0mm}
\end{figure*}

\begin{center}
    \begin{table*}
        \begin{center}
            \scalebox{0.75}{
            \begin{tabular}{c|c|ccc|ccc|cccc|cccc|c}
            \hline
            \multicolumn{2}{c|}{\multirow{2}{*}{Methods}} & \multicolumn{3}{c|}{View-point} & \multicolumn{3}{c|}{Scale} & \multicolumn{4}{c|}{Truncation} & \multicolumn{4}{c|}{Occlusion} & \multirow{2}{*}{All} \\
             \multicolumn{2}{c|}{ }  & easy & medi & hard & easy & medi & hard & none & src & tgt & both & none & src & tgt & both & \\
            \hline
            \hline
            \multicolumn{2}{c|}{Identity mapping} & 7.3 & 3.7 & 2.6 & 7.0 & 4.3 & 3.3 & 6.5 & 4.8 & 3.5 & 5.0 & 6.1 & 4.0 & 5.1 & 4.6 & 5.6 \\
            % \cline{1-19}
            \hline
            \multirow{4}{*}{\shortstack[1]{Transferred \\ \\ models}}
            & CNNGeo$_\textrm{res101}$~\cite{Rocco17}                & 25.2 & 10.7 & 5.9 & 22.3 & 16.1 & 8.5 & 21.1 & 12.7 & 15.6 & 13.9 & 20.0 & 14.9 & 14.3 & 12.4 & 18.1 \\
            & A2Net$_\textrm{res101}$~\cite{paul2018attentive}       & 27.5 & 12.4 & 6.9 & 24.1 & 18.5 & 10.3 & 22.9 & 15.2 & 17.6 & 15.7 & 22.3 & 16.5 & 15.2 & 14.5 & 20.1 \\ 
            & WeakAlign$_\textrm{res101}$~\cite{Rocco18}             & 29.4 & 12.2 & 6.9 & 25.4 & 19.4 & 10.3 & 24.1 & 16.0 & 18.5 & 15.7 & 23.4 & 16.7 & 16.7 & 14.8 & 21.1  \\ 
            % & NC-Net$_\textrm{res101}$~\cite{rocco2018neighbourhood} & 34.0 & 18.6 & 12.8 & 31.7 & 23.8 & 14.2 & 29.1 & 22.9 & 23.4 & 21.0 & 29.0 & 21.1 & 21.8 & 19.6 & 26.4  \\ 
            & NC-Net$_\textrm{res101}$~\cite{rocco2018neighbourhood} & \underline{34.0} & \underline{18.6} & \underline{12.8} & \underline{31.7} & \underline{23.8} & \underline{14.2} & \underline{29.1} & \underline{22.9} & \underline{23.4} & \underline{21.0} & \underline{29.0} & \underline{21.1} & \underline{21.8} & \underline{19.6} & \underline{26.4}  \\ 
            % \cline{2-19}
            % & HPF$_{\mathrm{res101}}$ (Ours) &  0.384 & 0.175 & 0.121 & 0.356 & 0.235 & 0.149 & 0.322 & 0.223 & 0.22 & 0.318 & 0.317 & 0.217 & 0.207 & 0.333 & 0.26 & 1.0  & 0.288 \\
            \hline
            % \hline
            \multirow{4}{*}{\shortstack[1]{SPair-71k \\ \\ trained \\ \\  models}}
            & CNNGeo$_\textrm{res101}$~\cite{Rocco17}                & 28.8 & 12.0 & 6.4 & 24.8 & 18.7 & 10.6 & 23.7 & 15.5 & 17.9 & 15.3 & 22.9 & 16.1 & 16.4 & 14.4 & 20.6 \\
            & A2Net$_\textrm{res101}$~\cite{paul2018attentive}       & 30.9 & 13.3 & 7.4 & 26.1 & 21.1 & 12.4 & 25.0 & 17.4 & 20.5 & 17.6 & 24.6 & 18.6 & 17.2 & 16.4 & 22.3 \\ 
            & WeakAlign$_\textrm{res101}$~\cite{Rocco18}             & 29.3 & 11.9 & 7.0 & 25.1 & 19.1 & 11.0 & 24.0 & 15.8 & 18.4 & 15.6 & 23.3 & 16.1 & 16.4 & 15.7 & 20.9 \\ 
            & NC-Net$_\textrm{res101}$~\cite{rocco2018neighbourhood} & 26.1 & 13.5 & 10.1 & 24.7 & 17.5 & 9.9 & 22.2 & 17.1 & 17.5 & 16.8 & 22.0 & 16.3 & 16.3 & 15.2 & 20.1 \\ 
            % \cline{2-19}
            \hline
            \multicolumn{2}{c|}{HPF$_{\mathrm{res50}}$ (ours)} & \underline{35.0} & \underline{18.9} & \underline{13.6} & \underline{32.0} & \underline{25.1} & \underline{15.4} & \underline{29.7} & \underline{24.5} & \underline{23.5} & \underline{22.9} & \underline{29.6} & \underline{22.9} & \underline{22.1} & \underline{21.3} & \underline{27.2} \\
            \multicolumn{2}{c|}{HPF$_{\mathrm{res101}}$ (ours)} & \textbf{35.6} & \textbf{20.3} & \textbf{15.5} & \textbf{33.0} & \textbf{26.1} & \textbf{15.8} & \textbf{31.0} & \textbf{24.6} & \textbf{24.0} & \textbf{23.7} & \textbf{30.8} & \textbf{23.5} & \textbf{22.8} &  \textbf{21.8} & \textbf{28.2}   \\
            \hline
            \end{tabular}}
        \vspace{-2.0mm}
        \caption{\label{tab:HPFanalysesTable} PCK analysis on SPair-71k. Difficulty levels of view points and scales are labeled easy, medium, and hard, while those of truncation and occlusion are indicated by none, source, target, and both.} 
        % the Result of weakalgin is lower than GeoCNN, because weakalign model overfit to the PF-PASCAL training set.
        % Note that weakalign [37] and ncnet model has lower score tooverfit PF-PASCAL training dataset.
        \vspace{-4.0mm}
        \end{center}
    \end{table*}
\end{center}

\smallbreak
\noindent \textbf{Comparison to proposal flow approach}~\cite{ham2016proposal}. The core differences between hyperpixel flow and proposal flow~\cite{ham2016proposal} are the changes in (1) matching primitives, from per-proposal geometric descriptor to hyperpixels, in order to handle problems of local-ambiguity and (2) matching algorithms, from PHM to RHM, in order to leverage hyperpixel geometry for efficiency. In Table~\ref{tab:stdBenchmarkTable}, significant performance improvements on three different benchmarks demonstrate that our features encoding high-level semantics while being agnostic to instance-specific details are crucial to establish robust correspondences. In addition, as shown in Table~\ref{tab:ablationSpeed}, the proposed voting method, RHM, with hyperpixels shows an impressive improvement in speed compared to~\cite{ham2016proposal}.
 
\smallbreak
\noindent \textbf{Inference time comparison.} With RHM, predicting dense correspondences for a single pair of images turns out to be much faster compared to other recent models. Table~\ref{tab:ablationSpeed} demonstrates the comparison of per-pair inference time on PF-PASCAL. While having more than 5\% improvements over current state-of-the-art approach~\cite{rocco2018neighbourhood}, the proposed model runs 4 to 13 times faster. With a slight trade-off on performance, hyperpixels with fewer layers and larger receptive field sizes enables real-time matching.

\smallbreak
\noindent \textbf{Ablation studies on matching.} 
To analyze the effects of RHM and its exponent factor $d$ in similarity $p(m_a)$, we experiment with replacing RHM with na\"ive nearest neighbor matching (NN) and also varying exponent $d$ of similarity. As shown in Table~\ref{tab:ablationModule}, the significant PCK gap between NN and RHM demonstrates the effectiveness of geometry matching. The performance improvement with $d \geq 2$ shows its effect of suppressing noisy votes in RHM.

\smallbreak
\noindent \textbf{Model analyses on SPair-71k benchmark.} We evaluate several recent methods~\cite{Rocco17, Rocco18, rocco2018neighbourhood, paul2018attentive} on our new benchmark dataset. In this experiment, our method tuned using the validation split of SPair-71k is evaluated on the test split of SPair-71k. For each method in comparison, we run two versions of each model: a trained model provided by the authors and the other further finetuned by ourselves on SPair-71k training set. The results are shown in Table~\ref{tab:hpftable}. We fail to successfully train the method of~\cite{Rocco18,rocco2018neighbourhood} on SPair-71k so that their performances drop when trained. We guess that their original learning objectives for weakly-supervised learning is fragile in presence of large view-point differences as in SPair-71k. We leave this issue for further investigation and will update the results at our benchmark page. 

%\jmlee{As shown in Table~\ref{tab:hpftable}, our model selected on the SPair-71k validation set is the best overall performance.
%NC-Net~\cite{rocco2018neighbourhood} works better at the rigid object classes such as bus, train, and TV, because the 4D convolution works as an a spatial regularizer like geometric transformation parameters.
%However, performances of NC-Net~\cite{rocco2018neighbourhood} and Weakalign~\cite{Rocco18} models decrease when they are trained on SPair-71k, because SPair-71k have large deformation, viewpoint differences, occlusion.
%Therefore, we need a flexible model to grasp more various patterns in semantic correspondence.}

SPair-71k has several annotation types such as view-point, scale, truncation and occlusion differences. In-depth analyses of each model using these annotations are summarized in Table~\ref{tab:HPFanalysesTable}. All models perform better with pairs of small differences, and view-point and scale differences significantly affect the performances. Yet, our method shows more robust results in terms of those variations compared to the others.
Figure~\ref{fig:figure4} shows some examples where our method finds reliable correspondences even under a large view-point and scale difference.

\section{Conclusion}
We have proposed a fast yet effective semantic matching method, hyperpixel flow, which leverages an optimized set of convolutional layer features pre-trained on a classification task. The impressive performance of the proposed method, which is only tuned with a small vadidation split without any end-to-end training, indicates that using relevant levels of multiple neural features is crucial in semantic correspondence. We believe further research in this direction is needed together with feature learning.  
To this end, we have also introduced a large-scale dataset, SPair-71k, with richer annotations for in-depth analyses, which is intended to resolve drawbacks of existing semantic correspondence datasets and to serve for supervised end-to-end learning of semantic correspondence.  
%For the details, see our supplementary.   

% Our dataset and code will be made available online. 

\smallbreak
\noindent \textbf{Acknowledgements.}
This work is supported by Samsung Advanced Institute of Technology (SAIT) and Basic Science Research Program (NRF-2017R1E1A1A01077999), and also in part by the Inria/NYU collaboration and the Louis Vuitton/ENS chair on artificial intelligence.

{\small
\bibliographystyle{ieee_fullname}
\bibliography{egbib}

\begin{thebibliography}{10}\itemsep=-1pt

\bibitem{bristow2015dense}
Hilton Bristow, Jack Valmadre, and Simon Lucey.
\newblock Dense semantic correspondence where every pixel is a classifier.
\newblock In {\em Proc. IEEE International Conference on Computer Vision
  (ICCV)}, 2015.

\bibitem{chen_cvpr14}
Xianjie Chen, Roozbeh Mottaghi, Xiaobai Liu, Sanja Fidler, Raquel Urtasun, and
  Alan Yuille.
\newblock Detect what you can: Detecting and representing objects using
  holistic models and body parts.
\newblock In {\em Proc. IEEE Conference on Computer Vision and Pattern
  Recognition (CVPR)}, 2014.

\bibitem{cho2013learning}
Minsu Cho, Karteek Alahari, and Jean Ponce.
\newblock Learning graphs to match.
\newblock In {\em Proc. IEEE International Conference on Computer Vision
  (ICCV)}, 2013.

\bibitem{cho2015unsupervised}
Minsu Cho, Suha Kwak, Cordelia Schmid, and Jean Ponce.
\newblock Unsupervised object discovery and localization in the wild:
  Part-based matching with bottom-up region proposals.
\newblock In {\em Proc. IEEE Conference on Computer Vision and Pattern
  Recognition (CVPR)}, 2015.

\bibitem{choy2016universal}
Christopher~B Choy, JunYoung Gwak, Silvio Savarese, and Manmohan Chandraker.
\newblock Universal correspondence network.
\newblock In {\em Proc. Neural Information Processing Systems (NeurIPS)}, pages
  2414--2422, 2016.

\bibitem{dalal2005histograms}
Navneet Dalal and Bill Triggs.
\newblock Histograms of oriented gradients for human detection.
\newblock In {\em Proc. IEEE Conference on Computer Vision and Pattern
  Recognition (CVPR)}, pages 886--893, 2005.

\bibitem{deng2009imagenet}
Jia Deng, Wei Dong, Richard Socher, Li-Jia Li, Kai Li, and Li Fei-Fei.
\newblock Imagenet: A large-scale hierarchical image database.
\newblock In {\em Proc. IEEE Conference on Computer Vision and Pattern
  Recognition (CVPR)}, pages 248--255, 2009.

\bibitem{donato2002approximate}
Gianluca Donato and Serge Belongie.
\newblock Approximate thin plate spline mappings.
\newblock In {\em Proc. European Conference on Computer Vision (ECCV)}, 2002.

\bibitem{everingham2015pascal}
Mark Everingham, S.~M.~Ali Eslami, Luc Van~Gool, Christopher K.~I. Williams,
  John Winn, and Andrew Zisserman.
\newblock The pascal visual object classes challenge: A retrospective.
\newblock {\em International Journal of Computer Vision (IJCV)},
  111(1):98--136, Jan 2015.

\bibitem{fathy2018hierarchical}
Mohammed~E Fathy, Quoc-Huy Tran, M Zeeshan~Zia, Paul Vernaza, and Manmohan
  Chandraker.
\newblock Hierarchical metric learning and matching for 2d and 3d geometric
  correspondences.
\newblock In {\em Proc. European Conference on Computer Vision (ECCV)}, pages
  803--819, 2018.

\bibitem{fei2004learning}
Li Fei-Fei, Rob Fergus, and Pietro Perona.
\newblock Learning generative visual models from few training examples: An
  incremental bayesian approach tested on 101 object categories.
\newblock In {\em Proc. IEEE Conference on Computer Vision and Pattern
  Recognition Workshops (CVPRW)}, page 178, 2004.

\bibitem{girshick2015deformable}
Ross Girshick, Forrest Iandola, Trevor Darrell, and Jitendra Malik.
\newblock Deformable part models are convolutional neural networks.
\newblock In {\em Proc. IEEE Conference on Computer Vision and Pattern
  Recognition (CVPR)}, pages 437--446, 2015.

\bibitem{griffin2007caltech}
Gregory Griffin, Alex Holub, and Pietro Perona.
\newblock Caltech-256 object category dataset.
\newblock {\em CalTech Report}, 2007.

\bibitem{ham2016proposal}
Bumsub Ham, Minsu Cho, Cordelia Schmid, and Jean Ponce.
\newblock Proposal flow.
\newblock In {\em Proc. IEEE Conference on Computer Vision and Pattern
  Recognition (CVPR)}, pages 3475--3484, 2016.

\bibitem{ham2018proposal}
Bumsub Ham, Minsu Cho, Cordelia Schmid, and Jean Ponce.
\newblock Proposal flow: Semantic correspondences from object proposals.
\newblock {\em IEEE Transactions on Pattern Analysis and Machine Intelligence
  (TPAMI)}, 40(7):1711--1725, 2018.

\bibitem{han2017scnet}
Kai Han, Rafael~S Rezende, Bumsub Ham, Kwan-Yee~K Wong, Minsu Cho, Cordelia
  Schmid, and Jean Ponce.
\newblock Scnet: Learning semantic correspondence.
\newblock In {\em Proc. IEEE International Conference on Computer Vision
  (ICCV)}, 2017.

\bibitem{hariharan2011semantic}
Bharath Hariharan, Pablo Arbel{\'a}ez, Lubomir Bourdev, Subhransu Maji, and
  Jitendra Malik.
\newblock Semantic contours from inverse detectors.
\newblock In {\em Proc. IEEE International Conference on Computer Vision
  (ICCV)}, 2011.

\bibitem{hariharan2015hypercolumns}
Bharath Hariharan, Pablo Arbel{\'a}ez, Ross Girshick, and Jitendra Malik.
\newblock Hypercolumns for object segmentation and fine-grained localization.
\newblock In {\em Proc. IEEE Conference on Computer Vision and Pattern
  Recognition (CVPR)}, pages 447--456, 2015.

\bibitem{he2016deep}
Kaiming He, Xiangyu Zhang, Shaoqing Ren, and Jian Sun.
\newblock Deep residual learning for image recognition.
\newblock In {\em Proc. IEEE Conference on Computer Vision and Pattern
  Recognition (CVPR)}, pages 770--778, 2016.

\bibitem{jeon2018parn}
Sangryul Jeon, Seungryong Kim, Dongbo Min, and Kwanghoon Sohn.
\newblock Parn: Pyramidal affine regression networks for dense semantic
  correspondence.
\newblock In {\em Proc. European Conference on Computer Vision (ECCV)}, pages
  351--366, 2018.

\bibitem{kanazawa2016warpnet}
Angjoo Kanazawa, David~W Jacobs, and Manmohan Chandraker.
\newblock Warpnet: Weakly supervised matching for single-view reconstruction.
\newblock In {\em Proc. IEEE Conference on Computer Vision and Pattern
  Recognition (CVPR)}, pages 3253--3261, 2016.

\bibitem{kim2013deformable}
Jaechul Kim, Ce Liu, Fei Sha, and Kristen Grauman.
\newblock Deformable spatial pyramid matching for fast dense correspondences.
\newblock In {\em Proc. IEEE Conference on Computer Vision and Pattern
  Recognition (CVPR)}, pages 2307--2314, 2013.

\bibitem{NIPS2018_7851}
Seungryong Kim, Stephen Lin, Sangryul Jeon, Dongbo Min, and Kwanghoon Sohn.
\newblock Recurrent transformer networks for semantic correspondence.
\newblock In {\em Proc. Neural Information Processing Systems (NeurIPS)}, 2018.

\bibitem{kim2017fcss}
Seungryong Kim, Dongbo Min, Bumsub Ham, Sangryul Jeon, Stephen Lin, and
  Kwanghoon Sohn.
\newblock Fcss: Fully convolutional self-similarity for dense semantic
  correspondence.
\newblock In {\em Proc. IEEE Conference on Computer Vision and Pattern
  Recognition (CVPR)}, pages 6560--6569, 2017.

\bibitem{kim2017dctm}
Seungryong Kim, Dongbo Min, Stephen Lin, and Kwanghoon Sohn.
\newblock Dctm: Discrete-continuous transformation matching for semantic flow.
\newblock In {\em Proc. IEEE International Conference on Computer Vision
  (ICCV)}, volume~6, 2017.

\bibitem{kong2016hypernet}
Tao Kong, Anbang Yao, Yurong Chen, and Fuchun Sun.
\newblock Hypernet: Towards accurate region proposal generation and joint
  object detection.
\newblock In {\em Proc. IEEE Conference on Computer Vision and Pattern
  Recognition (CVPR)}, pages 845--853, 2016.

\bibitem{krizhevsky2012imagenet}
Alex Krizhevsky, Ilya Sutskever, and Geoffrey~E Hinton.
\newblock Imagenet classification with deep convolutional neural networks.
\newblock In {\em Proc. Neural Information Processing Systems (NeurIPS)}, pages
  1097--1105, 2012.

\bibitem{laskar2019semantic}
Zakaria Laskar, Hamed~R Tavakoli, and Juho Kannala.
\newblock Semantic matching by weakly supervised 2d point set registration.
\newblock {\em arXiv preprint arXiv:1901.08341}, 2019.

\bibitem{li2006one}
Fei-Fei Li, Rob Fergus, and Pietro Perona.
\newblock One-shot learning of object categories.
\newblock {\em IEEE Transactions on Pattern Analysis and Machine Intelligence
  (TPAMI)}, 28(4):594--611, 2006.

\bibitem{lin2017feature}
Tsung-Yi Lin, Piotr Doll{\'a}r, Ross Girshick, Kaiming He, Bharath Hariharan,
  and Serge Belongie.
\newblock Feature pyramid networks for object detection.
\newblock In {\em Proc. IEEE Conference on Computer Vision and Pattern
  Recognition (CVPR)}, pages 2117--2125, 2017.

\bibitem{lin2014jointly}
Yen-Liang Lin, Vlad~I Morariu, Winston Hsu, and Larry~S Davis.
\newblock Jointly optimizing 3d model fitting and fine-grained classification.
\newblock In {\em Proc. European Conference on Computer Vision (ECCV)}, pages
  466--480, 2014.

\bibitem{liu2016sift}
Ce Liu, Jenny Yuen, and Antonio Torralba.
\newblock Sift flow: Dense correspondence across scenes and its applications.
\newblock In {\em Proc. European Conference on Computer Vision (ECCV)}, 2008.

\bibitem{liu2018darts}
Hanxiao Liu, Karen Simonyan, and Yiming Yang.
\newblock {DARTS}: Differentiable architecture search.
\newblock In {\em Proc. International Conference on Learning Representations
  (ICLR)}, 2019.

\bibitem{long2014convnets}
Jonathan~L Long, Ning Zhang, and Trevor Darrell.
\newblock Do convnets learn correspondence?
\newblock In {\em Proc. Neural Information Processing Systems (NeurIPS)}, pages
  1601--1609, 2014.

\bibitem{manen2013prime}
Santiago Manen, Matthieu Guillaumin, and Luc Van~Gool.
\newblock Prime object proposals with randomized prim's algorithm.
\newblock In {\em Proc. IEEE International Conference on Computer Vision
  (ICCV)}, pages 2536--2543, 2013.

\bibitem{beamsearch}
M.F. Medress, F.S. Cooper, J.W. Forgie, C.C. Green, D.H. Klatt, M.H. O'Malley,
  E.P. Neuburg, A. Newell, D.R. Reddy, B. Ritea, J.E. Shoup-Hummel, D.E.
  Walker, and W.A. Woods.
\newblock Speech understanding systems: Report of a steering committee.
\newblock {\em Artificial Intelligence}, 9(3):307 -- 316, 1977.

\bibitem{novotny2018self}
David Novotny, Samuel Albanie, Diane Larlus, and Andrea Vedaldi.
\newblock Self-supervised learning of geometrically stable features through
  probabilistic introspection.
\newblock In {\em Proc. IEEE Conference on Computer Vision and Pattern
  Recognition (CVPR)}, pages 3637--3645, 2018.

\bibitem{novotny16i-have}
David Novotny, Diane Larlus, and Andrea Vedaldi.
\newblock I have seen enough: Transferring parts across categories.
\newblock In {\em Proc. British Machine Vision Conference (BMVC)}, 2016.

\bibitem{novotny2017anchornet}
David Novotny, Diane Larlus, and Andrea Vedaldi.
\newblock Anchornet: A weakly supervised network to learn geometry-sensitive
  features for semantic matching.
\newblock In {\em Proc. IEEE Conference on Computer Vision and Pattern
  Recognition (CVPR)}, pages 2867--2876, 2017.

\bibitem{pont2017multiscale}
Jordi Pont-Tuset, Pablo Arbelaez, Jonathan~T Barron, Ferran Marques, and
  Jitendra Malik.
\newblock Multiscale combinatorial grouping for image segmentation and object
  proposal generation.
\newblock {\em IEEE Transactions on Pattern Analysis and Machine Intelligence
  (TPAMI)}, 39(1):128--140, 2017.

\bibitem{Rocco17}
Ignacio Rocco, Relja Arandjelovic, and Josef Sivic.
\newblock Convolutional neural network architecture for geometric matching.
\newblock In {\em Proc. IEEE Conference on Computer Vision and Pattern
  Recognition (CVPR)}, 2017.

\bibitem{Rocco18}
Ignacio Rocco, Relja Arandjelović, and Josef Sivic.
\newblock End-to-end weakly-supervised semantic alignment.
\newblock In {\em Proc. IEEE Conference on Computer Vision and Pattern
  Recognition (CVPR)}, 2018.

\bibitem{rocco2018neighbourhood}
Ignacio Rocco, Mircea Cimpoi, Relja Arandjelovi{\'c}, Akihiko Torii, Tomas
  Pajdla, and Josef Sivic.
\newblock Neighbourhood consensus networks.
\newblock In {\em Proc. Neural Information Processing Systems (NeurIPS)}, pages
  1656--1667, 2018.

\bibitem{rubinstein2013unsupervised}
Michael Rubinstein, Armand Joulin, Johannes Kopf, and Ce Liu.
\newblock Unsupervised joint object discovery and segmentation in internet
  images.
\newblock In {\em Proc. IEEE Conference on Computer Vision and Pattern
  Recognition (CVPR)}, pages 1939--1946, 2013.

\bibitem{paul2018attentive}
Paul~Hongsuck Seo, Jongmin Lee, Deunsol Jung, Bohyung Han, and Minsu Cho.
\newblock Attentive semantic alignment with offset-aware correlation kernels.
\newblock In {\em Proc. European Conference on Computer Vision (ECCV)}, 2018.

\bibitem{taniai2016joint}
Tatsunori Taniai, Sudipta~N Sinha, and Yoichi Sato.
\newblock Joint recovery of dense correspondence and cosegmentation in two
  images.
\newblock In {\em Proc. IEEE Conference on Computer Vision and Pattern
  Recognition (CVPR)}, pages 4246--4255, 2016.

\bibitem{ufer2017deep}
Nikolai Ufer and Bj{\"o}rn Ommer.
\newblock Deep semantic feature matching.
\newblock In {\em Proc. IEEE Conference on Computer Vision and Pattern
  Recognition (CVPR)}, pages 5929--5938, 2017.

\bibitem{uijlings2013selective}
Jasper~RR Uijlings, Koen~EA Van De~Sande, Theo Gevers, and Arnold~WM Smeulders.
\newblock Selective search for object recognition.
\newblock {\em International Journal of Computer Vision (IJCV)},
  104(2):154--171, 2013.

\bibitem{WahCUB_200_2011}
C. Wah, S. Branson, P. Welinder, P. Perona, and S. Belongie.
\newblock {The Caltech-UCSD Birds-200-2011 Dataset}.
\newblock Technical Report CNS-TR-2011-001, California Institute of Technology,
  2011.

\bibitem{Xiang2014BeyondPA}
Yu Xiang, Roozbeh Mottaghi, and Silvio Savarese.
\newblock Beyond pascal: A benchmark for 3d object detection in the wild.
\newblock {\em Proc. Winter Conference on Applications of Computer Vision
  (WACV)}, pages 75--82, 2014.

\bibitem{randomwire}
Saining Xie, Alexander Kirillov, Ross~B. Girshick, and Kaiming He.
\newblock Exploring randomly wired neural networks for image recognition.
\newblock {\em arXiv preprint arXiv:1904.01569}, 2019.

\bibitem{zhou2015flowweb}
Tinghui Zhou, Yong Jae~Lee, Stella~X Yu, and Alyosha~A Efros.
\newblock Flow{W}eb: Joint image set alignment by weaving consistent,
  pixel-wise correspondences.
\newblock In {\em Proc. IEEE Conference on Computer Vision and Pattern
  Recognition (CVPR)}, pages 1191--1200, 2015.

\bibitem{zoph2016neural}
Barret Zoph and Quoc~V Le.
\newblock Neural architecture search with reinforcement learning.
\newblock In {\em Proc. International Conference on Learning Representations
  (ICLR)}, 2017.

\bibitem{Zoph_2018_CVPR}
Barret Zoph, Vijay Vasudevan, Jonathon Shlens, and Quoc~V. Le.
\newblock Learning transferable architectures for scalable image recognition.
\newblock In {\em Proc. IEEE Conference on Computer Vision and Pattern
  Recognition (CVPR)}, June 2018.

\end{thebibliography}
}

\end{document}